\PassOptionsToPackage{table}{xcolor}

\documentclass[]{jingdong}

\usepackage[utf8]{inputenc}
\usepackage[T1]{fontenc}

\usepackage{amsfonts}
\usepackage{amsmath}
\usepackage{amssymb}
\usepackage{booktabs}
\usepackage{tabularx}
\usepackage{xcolor}
\usepackage{enumitem}
\usepackage{inconsolata}
\usepackage{listings}
\usepackage{nicefrac}
\usepackage{url}
\usepackage{xspace}
\usepackage{wrapfig}
\usepackage{booktabs}
\usepackage{pifont}
\newcommand{\cmark}{\ding{51}}
\newcommand{\xmark}{\ding{55}}

\crefformat{section}{\S#2#1#3}
\Crefformat{section}{\S#2#1#3}
\crefmultiformat{section}{\S#2#1#3}{ and \S#2#1#3}{, \S#2#1#3}{ and \S#2#1#3}
\Crefmultiformat{section}{\S#2#1#3}{ and \S#2#1#3}{, \S#2#1#3}{ and \S#2#1#3}
\crefrangeformat{section}{\S#3#1#4 to \S#5#2#6}
\Crefrangeformat{section}{\S#3#1#4 to \S#5#2#6}

\definecolor{templatekeyword}{HTML}{E1251B}
\definecolor{templatestring}{HTML}{0B7A75}
\definecolor{templatecomment}{HTML}{6B7280}
\definecolor{templatecodebg}{HTML}{F7F7F8}

\lstdefinestyle{templatecode}{
  basicstyle=\ttfamily\small,
  columns=fullflexible,
  backgroundcolor=\color{templatecodebg},
  frame=single,
  rulecolor=\color{black!15},
  numberstyle=\tiny\color{gray},
  keywordstyle=\color{templatekeyword},
  commentstyle=\color{templatecomment},
  stringstyle=\color{templatestring},
  showstringspaces=false,
  tabsize=2,
  breaklines=true,
  breakatwhitespace=true,
  captionpos=b,
  xleftmargin=3.4pt,
  xrightmargin=3.4pt
}

\usepackage{amsmath,amsfonts,bm}

\def\eqref#1{equation~\ref{#1}}
\def\1{\bm{1}}

\DeclareMathAlphabet{\mathsfit}{\encodingdefault}{\sfdefault}{m}{sl}
\SetMathAlphabet{\mathsfit}{bold}{\encodingdefault}{\sfdefault}{bx}{n}

\title{WorldBridge: Bridging Video and Action for Task Generalization in World Action Models}
\author[]{WorldBridge Team, Joy Future Academy}

\abstract{Achieving the broad task generalization of embodied models in physical environments remains a central challenge in robot learning.
We study this challenge in World Action Models (WAMs) through two complementary capabilities: video generalization, which predicts the visual evolution of new tasks, and action generalization, which translates these predictions into executable behavior through a video-to-action bridge.
By varying video and action supervision for tasks outside an initial training set, we examine their respective contributions to task transfer.
On RoboTwin 2.0, using only in-distribution video-action supervision yields a 25.07\% success rate on out-of-distribution tasks.
Introducing videos of these tasks without their action labels raises success to 55.90\%, while adding just one action-labeled demonstration per task further improves it to 70.70\%.
These findings suggest that broader video coverage enables task transfer by leveraging existing inverse-dynamics capabilities, while even sparse action supervision further strengthens those capabilities.
Guided by this perspective, we introduce \textbf{WorldBridge}, a WAM with separate video and action experts connected through structured attention and trained with joint flow matching.
WorldBridge incorporates visual memory at two temporal scales and action forcing, alongside a unified data processing pipeline that standardizes and filters heterogeneous robot trajectories and egocentric hand-derived supervision.
Across multiple manipulation benchmarks, WorldBridge achieves state-of-the-art performance among the compared methods, including 96.23\% on RoboTwin 2.0 full data setting and 80.9\% on clean to random setting, 82.8\% on RoboCasa GR1 TableTop, and 57.6\% on DOMINO Level-1.
Our study highlights the complementary roles of video prediction and action grounding in developing robots that generalize across tasks.\\
\\
\textbf{Project page:} \url{https://worldbridge-jd.github.io/}
}

\begin{document}

\maketitle

\section{Introduction}
\label{sec:introduction}

Task generalization marks a fundamental shift from earlier pretrained NLP models, such as BERT~\cite{devlin2019bert}, to modern large language models (LLMs)~\cite{brown2020language,yang2025qwen3,deepseekai2024v3}.
Whereas earlier models typically require task-specific fine-tuning, a single LLM can perform diverse tasks specified through natural-language instructions or a few in-context examples, without updating its parameters for each new task.
This capability has motivated the pursuit of generalist embodied agents that can translate human instructions into a broad repertoire of physical behaviors.
Vision-language-action (VLA) models represent an important step toward this goal, transferring knowledge from large-scale vision-language pretraining into robot control~\cite{brohan2023rt2, physicalintelligence2025pi05}.
However, understanding an instruction does not automatically provide the physical competence needed to execute it.
Bringing task generalization into robotics requires connecting semantic knowledge to the visual dynamics and motor behaviors of physical interaction.

Despite substantial progress, a prominent focus of embodied generalization research remains robustness to changes in the conditions under which a task is performed, including object arrangements, camera viewpoints, lighting, and visual appearance.
Benchmarks such as LIBERO-Plus systematically evaluate these variations, exposing the brittleness of otherwise successful policies~\cite{fei2026liberoplus}.
Task generalization introduces an additional challenge: a robot must execute behaviors beyond those covered by its original training tasks.
Achieving this capability requires understanding how a newly requested task should unfold and producing the actions that realize it, even when corresponding robot demonstrations are scarce or unavailable.

\begin{figure}[t]
    \centering
    \includegraphics[width=\textwidth]{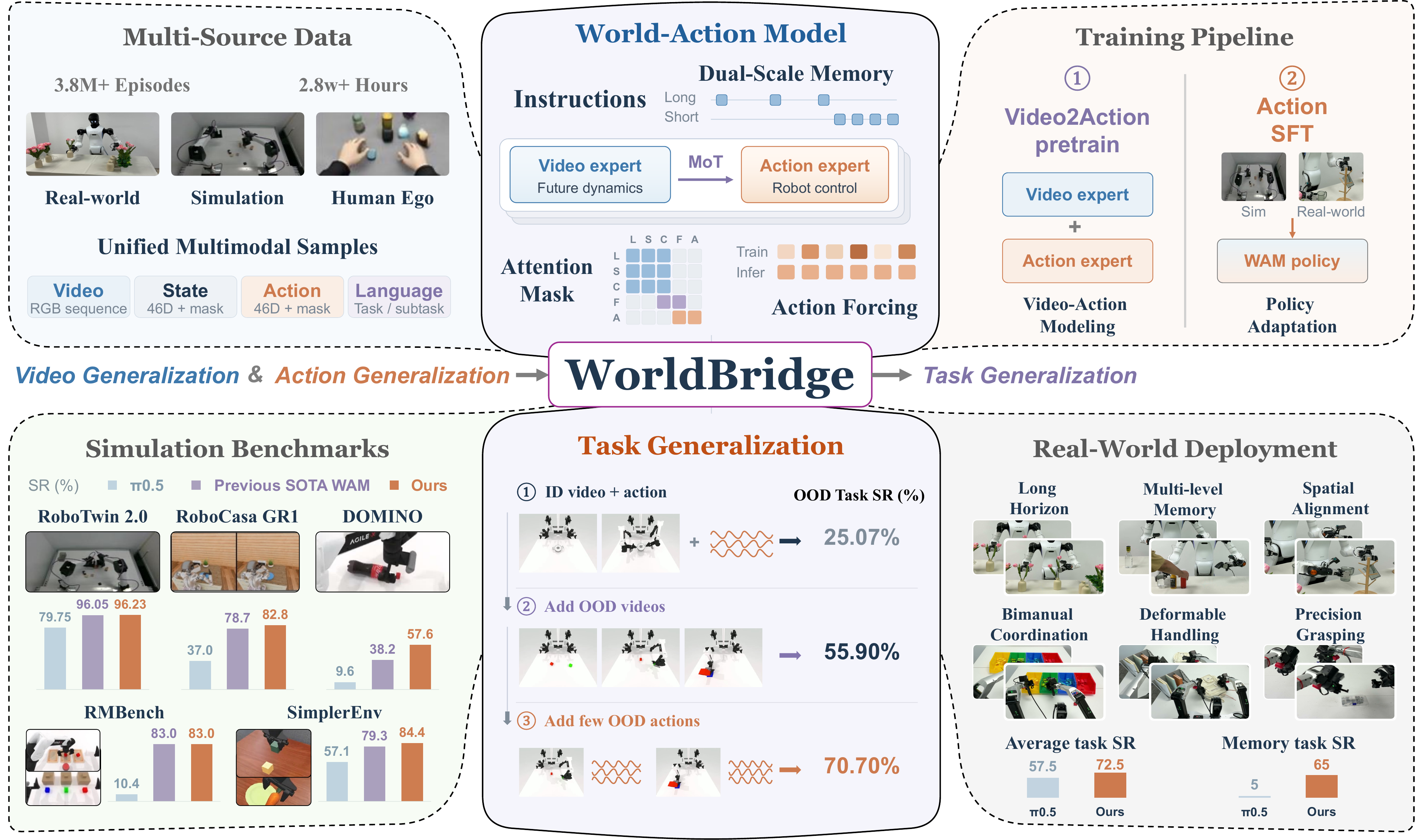}
    \caption{\textbf{Overview of our WorldBridge, a foundational WAM for task generalization.} WorldBridge utilizes heterogeneous data sources and is trained with a simple but effective model architecture in two stages. The bottom panels report task-generalization results under different video and action supervision settings, benchmark performance, and real-world deployment results.}
    \label{fig:teaser}
\end{figure}

World Action Models (WAMs) provide a promising foundation for addressing this challenge by coupling visual prediction with action generation.
Video offers dense supervision of physical interactions, allowing models to learn how scenes evolve beyond the coverage of action-labeled robot data.
DreamZero demonstrates that video priors can support unseen-task execution and transfer from video-only demonstrations~\cite{ye2026dreamzero}, while Zero-WAM explores task generalization through in-context human video guidance~\cite{zhou2026zerowam}.
Recent advances in native video-action pretraining~\cite{zhang2026native}, systematic WAM pretraining~\cite{wang2026openwam}, and temporal context~\cite{huang2026longwam} further strengthen this paradigm.
These developments motivate a more specific question: \emph{how do broader video coverage and additional action supervision contribute to task generalization, and when does the connection between them become the limiting factor?}

In this work, we investigate task generalization in WAMs through two complementary capabilities: {video generalization} and {action generalization}.
Video generalization enables a model to predict the visual evolution required by tasks beyond its initial training distribution.
Action generalization refers to the ability to translate those visual predictions into executable actions beyond the tasks that supplied action supervision.
This translation forms a \emph{video-to-action bridge}, serving the role of an inverse dynamics model (IDM).
A model may predict an appropriate visual future yet lack the action grounding needed to realize it.
Conversely, a transferable bridge can allow knowledge acquired from additional videos to improve task execution without corresponding action labels, which motivates studying the availability of video and action supervision separately.

\begin{figure}[t]
    \centering
    \includegraphics[width=\textwidth]{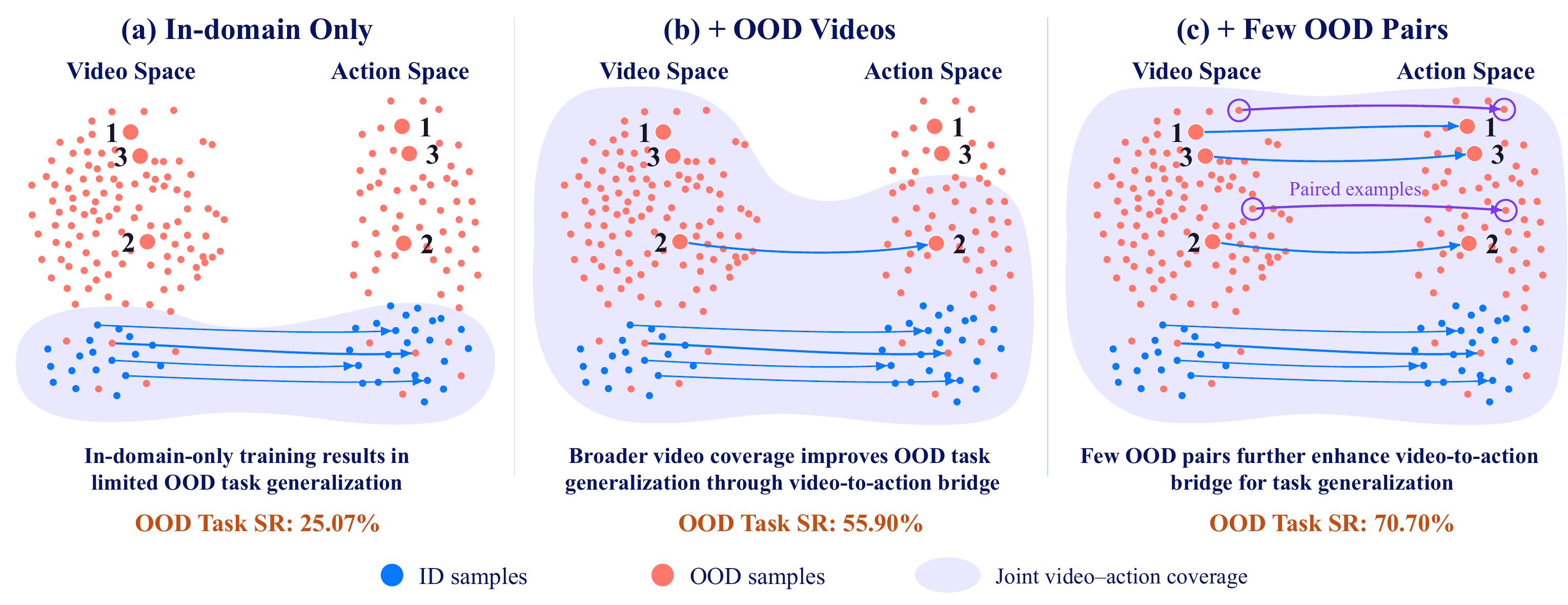}
    \caption{\textbf{Task generalization under progressively expanded supervision.} Compared with in-domain (ID) video-action training alone (25.07\% OOD success rate), adding out-of-distribution (OOD) videos increases success to 55.90\%, while adding a few paired OOD video-action examples further improves it to 70.70\%.}
    \label{fig:task_generalization}
\end{figure}

% Match these labels to the conceptual figure and supervision-ablation table
% in the main manuscript (Figure 1 and Table 2 in the supplied draft).
Figure~\ref{fig:task_generalization} illustrates this progression, which we examine by varying supervision for tasks outside an initial in-distribution (ID) training set.
With only ID video-action pairs, the model achieves nontrivial but limited generalization, obtaining a 25.07\% success rate on out-of-distribution (OOD) tasks.
Adding videos of those OOD tasks while keeping action supervision restricted to the ID set increases OOD success to 55.90\%.
This improvement suggests that expanded video coverage can support new task execution through a bridge learned from ID action supervision.
Keeping the video set fixed, adding only one action-labeled demonstration per OOD task further raises success to 70.70\%.
These rapid gains are consistent with sparse paired examples strengthening the bridge for visual transitions that are already covered by the video data.
Together, these observations support a complementary account of task generalization: video supervision expands the range of task evolutions a WAM can model, while action supervision improves its ability to turn that knowledge into reliable control.

Guided by this perspective, we introduce \textbf{WorldBridge}, a WAM that couples video prediction and action generation through a Mixture-of-Transformers (MoT) architecture~\cite{liang2024mot}.
WorldBridge maintains separate video and action experts trained with a joint flow-matching objective.
A structured attention mask establishes a directional information path from observed context to future video representations and then to action predictions, making predicted visual dynamics an explicit source of conditioning for control.
Short-term and long-term visual memories provide information about local motion and task progression.
We additionally employ action forcing, assigning independent noise levels to actions within a chunk during training so that the action expert learns to predict actions using temporal context at varying levels of corruption.

To support learning across heterogeneous sources, we develop a unified pretraining data pipeline for robot trajectories and egocentric human demonstrations.
The pipeline standardizes control semantics and maps available robot controls and human hand-derived targets into a shared 46-dimensional action representation.
Dimension-wise validity masks preserve partial supervision across embodiments with different control interfaces.
Robot trajectories undergo kinematic and temporal consistency checks, while egocentric hand trajectories are standardized in a common reference frame and filtered using geometric and motion constraints.
This processing supports broader visual coverage together with usable action supervision, providing a common interface for learning the video-to-action bridge across diverse data sources.

Our evaluations demonstrate strong performance across a range of manipulation tasks.
WorldBridge achieves state-of-the-art results among the compared methods on multiple benchmarks, with average success rates of 96.23\% on RoboTwin 2.0, 80.9\% in the RoboTwin 2.0 C2R setting, 82.8\% on RoboCasa GR1 TableTop, and 57.6\% on DOMINO Level-1.
On RMBench, it matches the best result in our comparison, with an 83.0\% success rate.
We further conduct real-world evaluations on the Agibot-G2 and AC-1 platforms, covering long-horizon execution, memory-dependent tasks, spatial localization, and bimanual coordination.
Together with our task-generalization study, these results support the value of jointly developing video prediction and action grounding.

Our contributions are threefold:
\begin{itemize}
    \item We study task generalization in WAMs through the complementary roles of video generalization and action generalization, quantifying the gains from additional task videos and sparse action supervision.
    \item We introduce WorldBridge and a unified data processing pipeline that connect visual dynamics to executable control across heterogeneous robot and egocentric demonstrations.
    \item We evaluate WorldBridge across RoboTwin 2.0, RoboCasa GR1 TableTop, RMBench, DOMINO Level-1, SimplerEnv-WidowX, and RoboDojo, covering diverse manipulation demands and distribution shifts. We further deploy it on AgiBot G2 and ARX AC-1, fine-tuning the same jointly pretrained checkpoint for each platform and evaluating long-horizon execution, memory-dependent interaction, spatial alignment, and bimanual coordination.
\end{itemize}

\begin{figure*}[t]
\centering
\includegraphics[width=\textwidth]{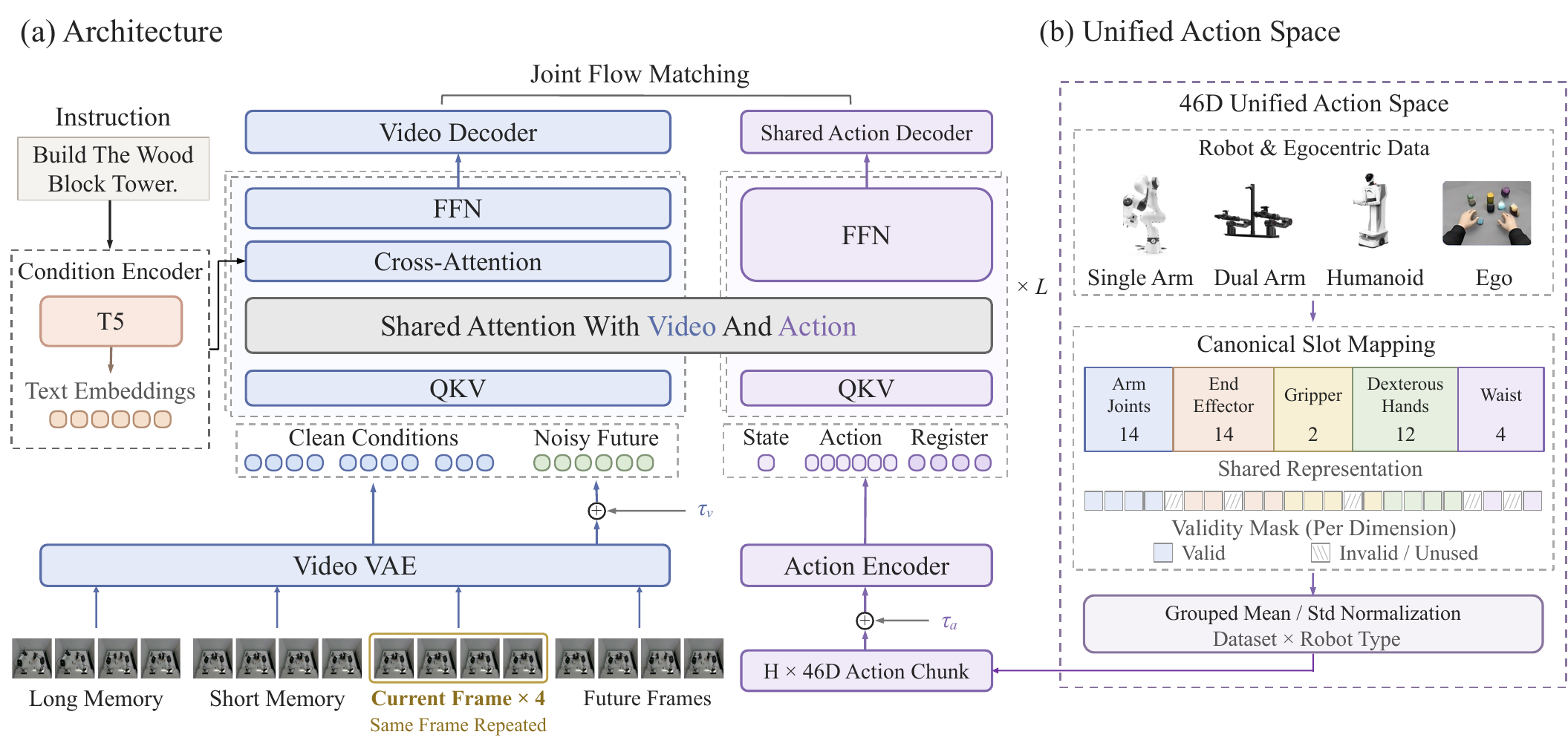}
\caption{\textbf{Overview of the WorldBridge model architecture.}}
\label{fig:model}
\end{figure*}
\section{Method}

\subsection{Model}

\subsubsection{Architecture}
\label{sec:architecture}

WorldBridge adopts a Mixture of Transformers (MoT) architecture comprising a video expert and an action expert. The two experts maintain modality-specific parameters and exchange information through joint self-attention, enabling coupled modeling of future visual observations and robot actions. Each expert contains its own feed-forward networks and output decoder, while the video expert additionally incorporates language features through cross-attention~\cite{wan2025wan}. This architecture preserves modality-specific processing while allowing visual dynamics and action generation to inform one another. An overview is shown in Fig.~\ref{fig:model}.

Given visual observations $\mathcal{O}$ and a language instruction
$\ell$, we construct the conditioning context $c$ using a video VAE
and a T5 text encoder, respectively.
\begin{equation}
    c = \left(
        \mathcal{E}_{\mathrm{VAE}}(\mathcal{O}),
        \mathcal{E}_{\mathrm{T5}}(\ell)
    \right).
\end{equation}
The clean prediction targets
consist of future video latents
$\mathbf{z} = \mathcal{E}_{\mathrm{VAE}}(\mathcal{V})$
and an action chunk
$\mathbf{A}_t = [\mathbf{a}_t,\ldots,\mathbf{a}_{t+H-1}]$,
where $t$ denotes the current environment timestep and $H$ is the
action prediction horizon.

We independently sample a video flow timestep $\tau_v$ and an action
flow timestep $\tau_a$ in $[0,1]$. Using a linear flow-matching path, we construct the noisy inputs as
\begin{equation}
\begin{aligned}
    \mathbf{z}(\tau_v)
    &= (1-\tau_v)\mathbf{z}
       + \tau_v\boldsymbol{\epsilon}_v, \\
    \mathbf{A}_t(\tau_a)
    &= (1-\tau_a)\mathbf{A}_t
       + \tau_a\boldsymbol{\epsilon}_a,
\end{aligned}
\end{equation}
where $\boldsymbol{\epsilon}_v$ and $\boldsymbol{\epsilon}_a$
are independent standard Gaussian noise tensors with shapes matching
their respective targets. Under this convention, $\tau=0$ corresponds
to clean data and $\tau=1$ to pure noise. Independent sampling of
$\tau_v$ and $\tau_a$ allows the two experts to process inputs at
different noise levels during training.

Conditioned on their respective flow timesteps, the two experts
process the noisy inputs through $L$ MoT blocks, exchanging information
through joint self-attention. The video and action decoders produce
the corresponding velocity predictions:
\begin{equation}
    \left(
        \hat{\mathbf{u}}_v,
        \hat{\mathbf{u}}_a
    \right)
    =
    f_{\theta}\left(
        \mathbf{z}(\tau_v),
        \mathbf{A}_t(\tau_a),
        \tau_v,
        \tau_a;
        c
    \right).
\end{equation}
The model is optimized through joint flow matching, with the predicted
velocities regressed toward
\begin{equation}
\begin{aligned}
    \mathbf{u}_v^{\star}
    &= \boldsymbol{\epsilon}_v - \mathbf{z}, \\
    \mathbf{u}_a^{\star}
    &= \boldsymbol{\epsilon}_a - \mathbf{A}_t.
\end{aligned}
\end{equation}

At inference, future video latents and the action chunk are
independently initialized from standard Gaussian noise:
\begin{equation}
    \mathbf{z}(1) \sim \mathcal{N}(\mathbf{0},\mathbf{I}),
    \qquad
    \mathbf{A}_t(1) \sim \mathcal{N}(\mathbf{0},\mathbf{I}).
\end{equation}
Both are progressively denoised by integrating the learned velocity
fields from $\tau=1$ toward $\tau=0$. A first-order update takes
the form
\begin{equation}
\begin{aligned}
    \mathbf{z}(\tau_v-\Delta\tau_v)
    &=
    \mathbf{z}(\tau_v)
    - \Delta\tau_v\,\hat{\mathbf{u}}_v, \\
    \mathbf{A}_t(\tau_a-\Delta\tau_a)
    &=
    \mathbf{A}_t(\tau_a)
    - \Delta\tau_a\,\hat{\mathbf{u}}_a,
\end{aligned}
\end{equation}
where the velocity predictions are evaluated at the current states
and flow timesteps. Throughout denoising, the conditioning context
$c$ remains fixed. The final video latents $\mathbf{z}(0)$ can be
reconstructed through the VAE decoder, while $\mathbf{A}_t(0)$
constitutes the predicted action chunk.

\subsubsection{Memory}
\label{sec:memory}

To provide temporal context beyond the current observation, WorldBridge
maintains visual memory at two temporal scales. Short-term memory
provides recent observations for capturing local motion, while
long-term memory provides a broader view of task progression.
Let $\mathbf{o}_t$ denote the visual observation at the current
environment timestep. We construct two chronologically ordered
observation sequences:
\begin{equation}
\begin{aligned}
    \mathcal{O}^{\mathrm{short}}_t
    &= [
        \mathbf{o}_{t-4\delta_s},
        \mathbf{o}_{t-3\delta_s},
        \mathbf{o}_{t-2\delta_s},
        \mathbf{o}_{t-\delta_s}
    ], \\
    \mathcal{O}^{\mathrm{long}}_t
    &= [
        \mathbf{o}_{t-4\delta_l},
        \mathbf{o}_{t-3\delta_l},
        \mathbf{o}_{t-2\delta_l},
        \mathbf{o}_{t-\delta_l}
    ],
\end{aligned}
\end{equation}
where $\delta_s$ and $\delta_l$ are the short-term and long-term
sampling strides. The long-term
stride corresponds to the temporal span of one predicted action
chunk. Observations preceding the episode boundary are replaced
with the first observation of that episode.

We encode both memory sequences together with the current
observation using the causal video VAE. To align with its temporal
compression factor of four, we repeat the current observation
four times and construct
\begin{equation}
    \mathcal{P}_t =
    [
        \mathbf{p},
        \mathcal{O}^{\mathrm{long}}_t,
        \mathcal{O}^{\mathrm{short}}_t,
        \operatorname{Repeat}_4(\mathbf{o}_t)
    ],
\end{equation}
where $\mathbf{p}$ is a zero-valued temporal padding frame.
After VAE encoding, we discard the leading latent associated
with this padding frame and retain three latent frames:
\begin{equation}
    [
        \mathbf{m}^{\mathrm{long}}_t,
        \mathbf{m}^{\mathrm{short}}_t,
        \mathbf{z}^{\mathrm{cur}}_t
    ]
    =
    \operatorname{DropFirst}
    \left(\mathcal{E}_{\mathrm{VAE}}(\mathcal{P}_t)\right).
\end{equation}
These latent frames provide a compact representation of the
long-term history, short-term history, and current observation.
They are spatially tokenized and prepended to the noisy future
video tokens.

\subsubsection{Action Forcing}
\label{sec:action_forcing}

When a single flow timestep is shared across an entire action chunk, all action tokens are corrupted to the same noise level. Action forcing instead exposes the action expert to chunks containing actions at different noise levels. Since action-token inputs can vary substantially in their noise levels and are difficult to model in isolation, this design encourages each token to rely more on contextual visual information.

Specifically, we extend the scalar action flow timestep
$\tau_a$ to a vector
$\boldsymbol{\tau}_a =
[\tau_{a,0},\ldots,\tau_{a,H-1}]$.
% We optionally group consecutive actions to share a flow
% timestep. 
For a group size $g$, we independently sample
a flow timestep $s_{a,k}$ for each group
$k=0,\ldots,\lceil H/g \rceil-1$ from the training
flow-matching schedule, independently of the video timestep
$\tau_v$. The timestep assigned to action $j$ is
\begin{equation}
    \tau_{a,j} = s_{a,\lfloor j/g \rfloor},
    \qquad j=0,\ldots,H-1.
\end{equation}
% The group size is benchmark-dependent, with $g=1$ recovering
% independent per-action timestep sampling.
For the $j$-th action in the chunk, the noisy input is
\begin{equation}
    \widetilde{\mathbf{a}}_{t+j}
    =
    (1-\tau_{a,j})\mathbf{a}_{t+j}
    + \tau_{a,j}\boldsymbol{\epsilon}_{a,j},
    \qquad j=0,\ldots,H-1,
\end{equation}
where the noise vectors
$\boldsymbol{\epsilon}_{a,j}\sim\mathcal{N}(\mathbf{0},\mathbf{I})$
are sampled independently, even when actions share a flow
timestep. Here, actions are represented in the normalized
unified action space.

Each action is encoded as one token and receives its
corresponding flow-time embedding. The action expert therefore
processes a mixture of relatively clean and highly corrupted
actions within the same chunk. Through attention within the
action branch, each prediction can use the other actions at
their respective noise levels, while also conditioning on
future video features. The prediction target remains the
flow-matching velocity:
\begin{equation}
    \mathbf{u}^{\star}_{a,j}
    =
    \boldsymbol{\epsilon}_{a,j}-\mathbf{a}_{t+j}.
\end{equation}

At inference, all action tokens are initialized with
independent standard Gaussian noise and denoised jointly
using a shared action flow timestep. Thus, varying noise
levels across actions or groups are only used during
training, while the complete action chunk is generated
through joint denoising at inference.

\subsubsection{Attention Mask}
\label{sec:attention_mask}

\setlength{\intextsep}{2pt}
\begin{wrapfigure}{r}{0.48\columnwidth}
\centering
\includegraphics[width=\linewidth]{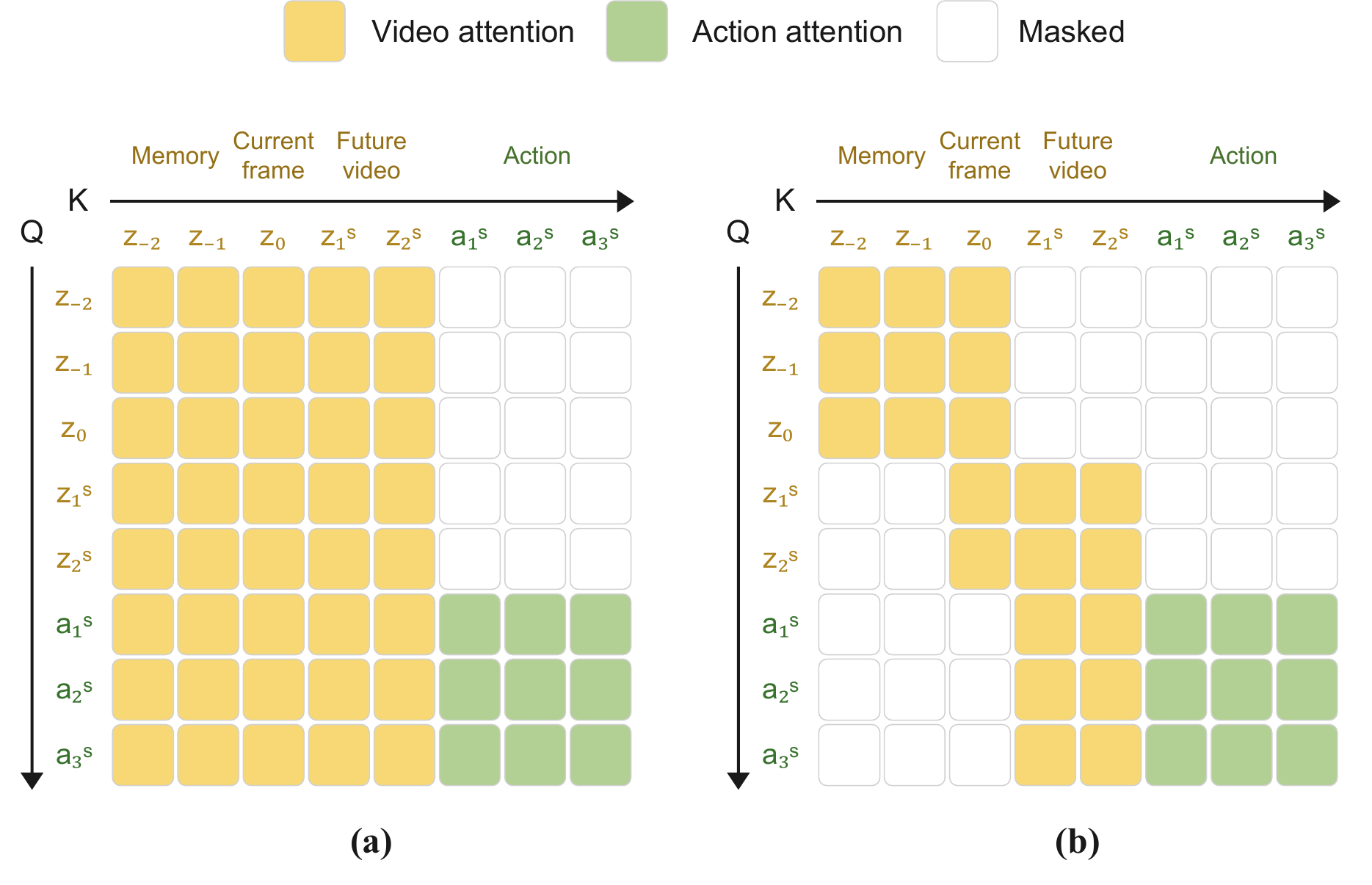}
\captionsetup{skip=2pt}
\caption{\textbf{Attention mask design.} (a) Attention mask for pretraining. (b) Attention mask for post-training.}
\label{fig:attention_mask}
\end{wrapfigure}

We adopt stage-specific attention masks to control information exchange
between video and action tokens within each MoT block.
As illustrated in Figure~\ref{fig:attention_mask}, the joint sequence
comprises four groups: memory tokens, current frame tokens, future frame tokens, and action tokens. We use a densely connected video mask during pretraining and introduce
more structured dependencies during post-training.

We use stage-specific attention masks to balance flexible multimodal learning during pretraining with an explicit video-to-action pathway during post-training. During pretraining, video tokens attend to all video tokens but not to action tokens, preserving action-independent video generation. Action tokens attend to both video and action tokens, allowing the action expert to learn from the full visual and action context. During post-training, memory and current-frame tokens attend to one another; future-video tokens attend to the current-frame and future-video tokens, and action tokens attend to future-video and action tokens. This structure routes information from memory and current observations through predicted future features to action predictions, encouraging actions to be grounded in the modeled visual evolution. For tasks with strong memory demands, we use the pretraining mask instead, allowing action tokens to access memory tokens directly.

\subsection{Data Processing}
\label{sec:data_processing}

\subsubsection{Robot Data Processing}

We curate heterogeneous robot demonstrations through a unified pipeline comprising semantic auditing, representation standardization, and trajectory filtering. We first map source-specific state and action fields into a shared representation with explicit validity masks. We then trim stationary episode boundaries, remove short trajectories, and apply kinematic and temporal consistency checks. Processing rules are configured for each source and embodiment to accommodate differences in annotation availability, control semantics, and signal characteristics.

\paragraph{Semantic Auditing and Representation Standardization.}
We first check what each source field represents, including joint order, EEF coordinates and rotations, gripper opening direction, and the meaning of recorded actions. State and action are mapped separately because a measured state and a control target are not interchangeable. Available values are placed in the shared 46-dimensional layout described in Section~\ref{sec:unified_action_space}. Separate state and action masks identify which entries are available; missing or non-finite values are set to zero and masked out.

EEF rotations are converted to unit quaternions in $(x,y,z,w)$ order. Gripper values are calibrated for each source so that $0$ means fully closed and $1$ means fully open. Each dexterous hand has six slots: bending of the little, ring, middle, index, and thumb fingers, followed by thumb rotation. Depending on the robot, we reorder the original channels, combine coupled finger joints, or map a shared hand-closure signal to the corresponding finger-bending slots. These slots therefore do not always represent independently measured finger joints. Unsupported hand motions remain masked. Four additional slots store supported waist and torso signals, such as waist pitch and torso lift, with a mapping specified for each robot; unused slots remain masked. EEF positions retain their original coordinate frames and units. Relative action targets and training normalization are defined separately in Section~\ref{sec:unified_action_space}.

\paragraph{Static Boundary Trimming and Length Filtering.}
We detect motion from adjacent state differences over dimensions that are valid and finite at both timesteps. A transition is considered active if any comparable dimension changes by more than $10^{-4}$ in the stored feature space. We retain the interval spanning the first and last active transitions, including their endpoint frames. This removes stationary prefixes and suffixes while preserving pauses within task execution. Fully stationary episodes with observable states are discarded, whereas episodes without comparable valid state dimensions are left unchanged by this check. We subsequently remove trajectories shorter than 48 frames and update episode and video-window metadata to reflect the retained intervals.

\paragraph{Kinematic Consistency Filtering.}
For sources with comparable joint and EEF annotations, we compute the EEF pose from recorded joint positions using forward kinematics (FK). We account for source-specific base frames and tool definitions before comparing the FK pose with the recorded pose. Let $(\mathbf{p}_{\mathrm{FK}},\mathbf{R}_{\mathrm{FK}})$ and $(\mathbf{p}_{\mathrm{data}},\mathbf{R}_{\mathrm{data}})$ denote these two poses at the same timestep. We measure their position and orientation errors as
\begin{equation}
e_{\mathrm{pos}}=\|\mathbf{p}_{\mathrm{FK}}-\mathbf{p}_{\mathrm{data}}\|_2,\qquad e_{\mathrm{rot}}=\arccos\!\left(\frac{\operatorname{tr}(\mathbf{R}_{\mathrm{FK}}^\top\mathbf{R}_{\mathrm{data}})-1}{2}\right).
\end{equation}
The argument of $\arccos$ is clipped to $[-1,1]$ for numerical stability. A frame passes when the applicable errors are below embodiment-specific thresholds, and an episode is retained when a sufficient fraction of evaluated frames passes. Invalid required poses count as failures, while position-only configurations omit the orientation check. For selected sources with an unknown fixed frame relationship, we estimate a single rigid transformation per episode before computing the errors. This alignment is used only for consistency assessment.

\paragraph{Signal Anomaly and State--Action Consistency Filtering.}
We apply three complementary signal checks and reject episodes that fail an applicable criterion.

\textit{Sudden-change detection.} We estimate a smooth trajectory using median and Savitzky--Golay filtering, then evaluate the residual together with second- and third-order temporal differences. A sample is flagged when an excessive residual coincides with an excessive higher-order difference. Robust, channel-specific thresholds distinguish abrupt anomalies from ordinary motion. Quaternion signs are aligned temporarily during detection to avoid artificial discontinuities between equivalent rotations. The smoothing and sign alignment are used for detection and do not replace the stored signals.

\textit{State--action consistency.} For comparable joint dimensions with sufficient motion, we compare the directions of changes in smoothed state and action signals. We search offsets within $\pm15$ frames and retain the best directional agreement for each eligible dimension. Episodes are rejected when the minimum agreement across eligible dimensions falls below the source-specific threshold. This procedure assesses temporal consistency without shifting the saved action sequences.

\textit{Extreme-value detection.} We estimate the 1st and 99th percentiles from valid samples within each repository. The admissible interval extends beyond these quantiles by the larger of five times their span and a channel-specific minimum margin. Episodes containing out-of-range or non-finite active values are rejected. Quaternion, gripper, and dexterous-hand channels are exempt from this range test. Source-specific configurations determine the applicable checks and exemptions.

\begin{figure*}[t]
\centering
\includegraphics[width=\textwidth]{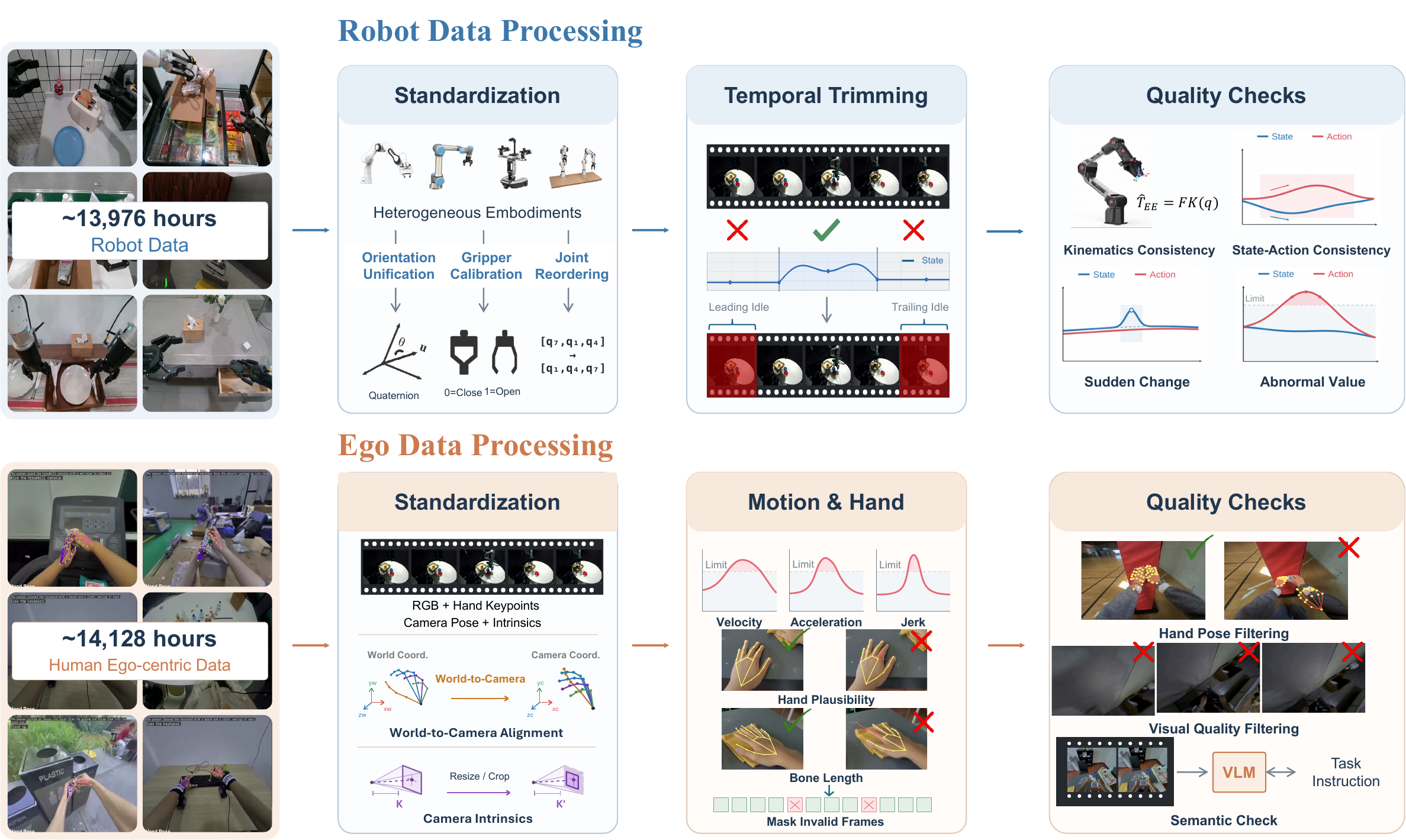}
\caption{Overview of the WorldBridge data processing pipeline.}
\label{fig:data_processing}
\end{figure*}

\subsubsection{Ego Data Processing}

We process egocentric demonstrations into synchronized visual
observations and hand-derived action targets. Each sequence
contains egocentric RGB frames, camera-to-world poses, and
21 three-dimensional keypoints per hand in a consistent
MANO ordering. Available task and subtask annotations provide
the language context. Our pipeline standardizes the coordinate
representation, maps hand geometry into the shared action
layout, and filters unreliable supervision using geometric
and temporal consistency checks.

\paragraph{Coordinate Standardization.}
To account for the moving egocentric camera, we express future
hand geometry in the camera frame of the current observation.
Let $\mathbf{T}^{w}_{c_t}\in SE(3)$ denote the camera-to-world
transformation at timestep $t$, and let
$\bar{\mathbf{x}}^{c_{t+k}}_{t+k,j}$ be the homogeneous
coordinate of hand keypoint $j$ observed at a future timestep.
Its coordinate in the current camera frame is
\begin{equation}
    \bar{\mathbf{x}}^{c_t}_{t+k,j}
    =
    \left(\mathbf{T}^{w}_{c_t}\right)^{-1}
    \mathbf{T}^{w}_{c_{t+k}}
    \bar{\mathbf{x}}^{c_{t+k}}_{t+k,j}.
\end{equation}
The reference frame remains fixed throughout the prediction
chunk, placing future hand trajectories in a common coordinate
system despite changes in camera pose. Video and geometry
are sampled on a shared physical-time schedule to preserve
their temporal correspondence.

\paragraph{Hand-to-Action Mapping.}
We derive a canonical wrist pose from each transformed hand.
The wrist keypoint defines its position, while the wrist and
metacarpophalangeal joints define its orientation. Specifically,
the forward axis points from the wrist toward the centroid
of the four finger bases, and a least-squares palm plane
provides the palm normal. A handedness correction gives the
left and right hands consistent axis semantics.

We additionally extract a continuous gripper-openness target
from the distance between the thumb tip and a virtual opposing
finger formed from the index and middle fingertips.
Normalizing this distance by palm width reduces dependence
on hand size. For dexterous-hand supervision, finger bending
and thumb opposition are mapped to six bounded joint-position
targets per hand.

\paragraph{Quality Filtering and Validity Masks.}
We assess hand annotations using biomechanical plausibility,
coordinate bounds, bone lengths, and temporal motion
statistics, including velocity, acceleration, and jerk.
Motion derivatives are computed using actual timestamps
and world-coordinate hand trajectories. Camera rotations
are checked for orthogonality and determinant consistency.
Dataset-specific thresholds accommodate differences in
motion statistics across sources.

Quality annotations are consumed at the sampled action
timesteps. Targets associated with rejected frames are
masked, while windows with more than 20\% rejected action
timesteps are discarded. An invalid reference camera pose
invalidates the corresponding action chunk. During action
construction, non-finite coordinates, padded frames, and
degenerate hand geometry are further excluded through
validity masks. Missing observations of one hand do not
automatically invalidate the other. This preserves usable
visual context and partial hand supervision without treating
unavailable action dimensions as valid zero-valued targets.

\subsection{Pretraining Recipe}
\label{sec:pretraining_recipe}

We pretrain WorldBridge jointly on robot demonstrations and egocentric
human demonstrations processed by the pipeline in
Section~\ref{sec:data_processing}. Below, we describe the training
corpus, the shared action representation, and the pretraining setup.

\subsubsection{Robot and Ego Data}
\label{sec:pretraining_data}

\paragraph{Robot Data.}
Our corpus combines real-world and simulated robot demonstrations. Real-world sources include ABC-130K~\cite{allshire2026abc}, AgiBotWorld-Beta~\cite{bu2025agibot}, AgiBotWorld2026~\cite{agibotworld2026}, DROID~\cite{khazatsky2024droid}, Galaxea Open-World~\cite{jiang2025galaxea}, RoboMIND~\cite{wu2024robomind}, RoboCOIN~\cite{wu2025robocoin}, and RoboMIND~2.0~\cite{hou2025robomind2}. Simulation sources include GenieSim~3.0~\cite{yin2026geniesim}, and InternData-A1~\cite{tian2025interndata}, RoboMIND~2.0 also contributes simulation subsets. Together, these datasets cover single-arm, bimanual, and humanoid platforms equipped with parallel-jaw grippers or dexterous hands. Table~\ref{tab:robot_ego_data_statistics} summarizes the curated sources.

\paragraph{Ego Data.}
Our egocentric corpus comprises EgoStandard \cite{lightwheel2026egosuite}, Lightwheel \cite{lightwheel2026egosuite}, XDOF,
CyberOrigin, GenRobot, Xperience \cite{xperience_10m}, and EgoDex \cite{hoque2026egodex}. These sources provide
first-person visual observations and hand-derived supervision,
which we convert into the shared action representation through the
ego processing pipeline. Validity masks retain available hand
supervision while excluding unreliable or missing targets.

\paragraph{Dataset Statistics.}
Table~\ref{tab:robot_ego_data_statistics} reports episode counts and durations before and after curation for the listed sources. Durations are computed using their specific frame rates, and retention is the ratio of output to input hours. The lower retention of ABC-130K and RoboMIND primarily reflects invalid or unavailable EEF supervision and embodiment-level subset selection.

After curation, the listed corpus contains 1,275,715 robot episodes
(9,729.58 hours) and 2,105,846 ego episodes (12,591.63 hours), totaling
3,381,561 episodes and 22,321.21 hours. The contribution of each source during training additionally depends
on the sampling mixture.

\begin{table*}[t]
\centering
\small
\setlength{\tabcolsep}{5pt}
\renewcommand{\arraystretch}{1.1}
\caption{Robot and ego corpus statistics before and after curation, excluding RoboCasa and RoboTwin. For robot data, hour retention includes subset selection, boundary trimming, length filtering, and quality filtering. For ego data, after-curation episode counts reflect episode-level annotation filtering, and hours count frames passing all seven configured quality checks within retained episodes. Durations are computed using repository-specific frame rates.}
\label{tab:robot_ego_data_statistics}
\begin{tabular*}{\textwidth}{@{\extracolsep{\fill}}lrrrrr@{}}
\toprule
& \multicolumn{2}{c}{Before curation}
& \multicolumn{2}{c}{After curation}
& Retention \\
\cmidrule(lr){2-3}
\cmidrule(lr){4-5}
Dataset & Episodes & Hours & Episodes & Hours & (\% hours) \\
\midrule
\multicolumn{6}{l}{\textit{Robot datasets}} \\
ABC-130K~\cite{allshire2026abc} & 130,919 & 4,130.93 & 66,426 & 1,642.30 & 39.76 \\
AgiBotWorld-Beta~\cite{bu2025agibot} & 158,776 & 2,392.17 & 155,587 & 2,286.54 & 95.58 \\
AgiBotWorld2026~\cite{agibotworld2026} & 22,436 & 456.85 & 18,233 & 390.66 & 85.51 \\
DROID~\cite{khazatsky2024droid} & 57,774 & 273.12 & 56,279 & 263.80 & 96.59 \\
Galaxea~\cite{jiang2025galaxea} & 18,983 & 454.66 & 17,883 & 423.37 & 93.12 \\
RoboCOIN~\cite{wu2025robocoin} & 98,580 & 688.84 & 94,877 & 653.23 & 94.83 \\
RoboMIND~\cite{wu2024robomind} & 73,882 & 218.31 & 25,171 & 36.91 & 16.91 \\
RoboMIND 2.0~\cite{hou2025robomind2} & 187,590 & 1,218.70 & 116,901 & 262.37 & 21.53 \\
GenieSim 3.0~\cite{yin2026geniesim} & 139,591 & 252.34 & 133,801 & 211.38 & 83.77 \\
InternData-A1~\cite{tian2025interndata} & 613,119 & 3,890.74 & 590,557 & 3,559.02 & 91.47 \\
\midrule
\textbf{Robot subtotal}
& \textbf{1,501,650} & \textbf{13,976.65}
& \textbf{1,275,715} & \textbf{9,729.58}
& \textbf{69.61} \\
\midrule
\multicolumn{6}{l}{\textit{Ego datasets}} \\
EgoStandard & 1,623,216 & 8,170.37 & 1,623,216 & 7,656.85 & 93.71 \\
Lightwheel  & 31,576 & 1,049.92 & 31,576 & 987.62 & 94.07 \\
XDOF        & 42,059 & 1,000.44 & 42,059 & 971.46 & 97.10 \\
CyberOrigin & 35,062 & 1,026.51 & 35,062 & 985.22 & 95.98 \\
Xperience   & 312,743 & 1,156.00 & 123,138 & 477.29 & 41.29 \\
EgoDex      & 314,839 & 724.84 & 222,982 & 531.15 & 73.28 \\
GenRobot    & 27,813 & 1,000.14 & 27,813 & 982.05 & 98.19 \\
\midrule
\textbf{Ego subtotal}
& \textbf{2,387,308} & \textbf{14,128.22}
& \textbf{2,105,846} & \textbf{12,591.63}
& \textbf{89.12} \\
\midrule
\textbf{Total}
& \textbf{3,888,958} & \textbf{28,104.87}
& \textbf{3,381,561} & \textbf{22,321.21}
& \textbf{79.42} \\
\bottomrule
\end{tabular*}

\end{table*}

\subsubsection{Unified Action Space}
\label{sec:unified_action_space}

To accommodate heterogeneous robot embodiments within a shared
action expert, we organize actions into a fixed 46-dimensional
representation comprising arm joints (14D), end-effector poses
(14D), grippers (2D), dexterous hands (12D), and waist controls (4D).
Each embodiment populates its available fields, accompanied by a
dimension-wise validity mask. This shared layout preserves
embodiment-specific control variables while allowing a common
prediction interface. Unavailable dimensions are zero-filled and
excluded from action supervision.

\paragraph{End-Effector-Relative Representation.}
While the standardized data stores end-effector poses as positions
and quaternions, we express their prediction targets relative to
the corresponding end-effector frame at the current timestep.
For each end effector, let $(\mathbf{p}_t,\mathbf{R}_t)$ denote
the reference pose in the current action record, and
$(\mathbf{p}_{t,k},\mathbf{R}_{t,k})$ the pose of the $k$-th
target in the prediction chunk. We compute
\begin{equation}
\begin{aligned}
    \Delta\mathbf{p}_{t,k}
    &= \mathbf{R}_t^\top
       (\mathbf{p}_{t,k}-\mathbf{p}_t), \\
    \Delta\boldsymbol{\phi}_{t,k}
    &= \operatorname{Log}
       (\mathbf{R}_t^\top\mathbf{R}_{t,k})^\vee ,
\end{aligned}
\end{equation}
where $\operatorname{Log}(\cdot)^\vee$ maps a rotation matrix
to its three-dimensional rotation vector in radians.
The reference frame remains fixed throughout the chunk.
Each original seven-dimensional pose slot therefore contains
three translation components, three rotation components, and
one masked padding dimension. The remaining action fields retain
their standardized values. This representation expresses
end-effector motion in a local frame while retaining the shared
46-dimensional layout.

\paragraph{Masked Normalization and Supervision.}
We normalize action targets using dimension-wise statistics
computed separately for each dataset--robot-type group.
For an action component $a_{t,k,d}$ with validity
$m_{t,k,d}\in\{0,1\}$, the normalized target is
\begin{equation}
    \widetilde{a}_{t,k,d} =
    \begin{cases}
    \displaystyle
    \frac{a_{t,k,d}-\mu_{g,d}}
         {\max(\sigma_{g,d},\epsilon)},
    & m_{t,k,d}=1, \\[6pt]
    0, & m_{t,k,d}=0,
    \end{cases}
\end{equation}
where $g$ identifies the dataset--robot-type group.
The statistics are computed in the same action representation
used for training. Validity masks also account for the reference
pose required by the relative transformation.
During flow matching, the masks are applied to the clean actions,
sampled noise, and velocity targets. The action loss is first
averaged over valid dimensions within each action token, and
then over tokens containing valid supervision.

\paragraph{State Conditioning.}
We additionally encode the current frame's absolute 46-dimensional
action record as a state token for the action expert.
This state token retains the absolute end-effector poses and is
normalized using separate current-frame statistics, with
unavailable dimensions set to zero. It provides the reference
configuration for predicting the relative action chunk and is
excluded from the action prediction loss.

\subsubsection{Pretraining Setup}
\label{sec:pretraining_setup}

% Configuration reference: mixv4_launch_20261019/launch_node.sh and
% launch_node0.log, checked against launch/qwen35_wan_pretrain.sh and
% the pretraining policy, scheduler, and dataset reader on 2026-10-23.
% The step count below is the configured budget, not verified completion.
\paragraph{Optimization.}
We jointly optimize the video and action experts with equally weighted
video and action flow-matching losses, using AdamW with
$\beta_1=0.9$, $\beta_2=0.95$, weight decay $0.01$, and gradient
clipping at a norm of $0.5$. Both experts use a base learning rate
of $10^{-4}$.
We pretrain for 60,000 optimizer steps, linearly increasing the learning rate from $10^{-10}$ to $9.9\times10^{-5}$ during the first 2,000 steps.
Training uses BF16 on 64 GPUs, with 12 samples per GPU
and two gradient-accumulation steps, giving an effective batch size
of $12\times64\times2=1,536$.

\paragraph{Sampling and Training Windows.}
We sample robot and ego data with probabilities $0.7$ and $0.3$,
respectively, and weight repositories within each source category
by usable physical duration. Each chunk contains $H=32$ action
targets and eight future video frames, with one video frame per
four action timesteps. Robot and ego action grids are sampled at
15\,Hz and 30\,Hz, respectively. For sources configured with three camera views, we spatially compose
synchronized main-camera and left/right wrist-camera images into a
single frame at each timestep. The main view occupies the top region,
while the two wrist views, each resized to half the main view's width
and height, are placed side by side below it, forming a T-shaped layout.
The composite frame is then bilinearly resized to $320\times384$
pixels (width $\times$ height) without added padding and encoded as
one video stream by the shared video VAE. Sources configured for a
single view use the selected camera image resized to the same resolution.
The same camera layout is applied to memory, current, and future frames. The current
absolute action record is encoded as a 46-dimensional state token;
action targets use the normalized unified representation.

\paragraph{Attention Mask and Noise Sampling.}
We use the pretraining mask in Fig.~\ref{fig:attention_mask} (a).
Video tokens attend to all video tokens and do not attend to the
action branch. Action-branch tokens attend to all video tokens and
to one another. Thus, video prediction is independent of action
inputs, while action prediction can use the full video context.
The video and action flow timesteps are sampled independently.
Action forcing is disabled in this pretraining configuration:
all actions within a chunk share one action flow timestep.

\paragraph{Memory Configuration.}
Both short-term and long-term visual memory are enabled, with four
observations at each scale. On the source-specific action time grid,
short-term observations have offsets $[-8,-6,-4,-2]$, while long-term
observations have offsets $[-128,-96,-64,-32]$ relative to the current
observation. Thus, $\delta_s=2$ and $\delta_l=H=32$ in
Section~\ref{sec:memory}. The causal video VAE compresses these
histories into two memory latent frames alongside the current-frame
latent. All three provide clean conditioning context, with flow
timestep zero, under the pretraining attention mask.

\subsection{Efficiency Implementation}
\label{sec:efficiency_implementation}

% We accelerate action inference through operator fusion and CUDA Graph replay. The action denoising computation is compiled into optimized GPU kernels, fusing compatible operations to reduce intermediate memory traffic and kernel launches. CUDA Graphs capture the resulting GPU execution sequence and replay it during subsequent denoising steps, reducing Python dispatch and CPU-side kernel-launch overhead.
% We also decouple video and action denoising. The video branch runs for a single step to construct per-layer key–value features, which are reused across all ten action denoising steps and refreshed for each new observation. This avoids repeatedly evaluating the video branch while refining actions. Each inference produces a chunk of 40 actions, amortizing observation-processing costs across the action sequence.

\paragraph{Inference acceleration.} We accelerate inference by decoupling video and action denoising. At deployment, a naive implementation performs $N$ denoising steps in both the video and action branches, typically with $N=10$. Multi-step denoising in the computationally intensive video branch substantially increases inference cost. For acceleration, we perform $n$ synchronized denoising steps that update both branches and construct per-layer video key--value features; the remaining $N-n$ action denoising steps reuse the features from the final synchronized step. The directional attention mask permits this reuse without feedback from action tokens to video tokens, although freezing early video features approximates joint denoising. Visual features are refreshed for every observation, and future-video RGB decoding is omitted. Action chunks, comprising 48 steps in the paper setting, amortize observation encoding across the predicted sequence.

We further accelerate action inference through operator fusion and CUDA Graph replay. The action denoising computation is compiled with \texttt{torch.compile} into optimized GPU kernels, fusing compatible operations to reduce intermediate memory traffic and kernel launches. CUDA Graphs capture the resulting GPU execution sequence and replay it during subsequent denoising steps, reducing Python dispatch and CPU-side kernel-launch overhead. The policy runs in BF16 under inference mode. We also reuse language context, scheduler metadata, and timestep-dependent AdaLN tensors, and evaluate QKV projection, KV-buffer, and fixed-shape attention optimizations. Compilation and graph capture are completed during warmup.

\paragraph{Asynchronous execution.} To maintain smooth robot motion despite model inference latency, we decouple policy inference from action execution in real-world experiments. The server computes the next action chunk on the GPU while the client continues executing the current chunk. The client prefetches new predictions based on the remaining execution time of its current plan, with at most one outstanding request to avoid accumulating stale observations. Each predicted chunk is timestamped relative to its input observation. Upon receipt, expired actions are discarded, while the remaining actions retain their original execution times. The client resamples the chunk once onto the 100\,Hz servo grid, and the C++ controller handles trajectory splicing and final amplitude limiting.
\section{Experiments}

\subsection{Task Generalization}
\label{sec:task_generalization}

We investigate how video and action supervision contribute to task generalization in WAMs.
Table~\ref{tab:ood_shot_results} compares different supervision settings on RoboTwin 2.0.
Across all settings, we keep the ID video and action data fixed, while varying the availability of OOD videos and action labels.
Here, ID and OOD refer to a fixed task partition defined before introducing additional supervision.

\textbf{Generalization from ID supervision alone.}
When trained with only ID video-action supervision, the model achieves a high ID success rate of 91.85\%, but its OOD success rate is limited to 25.07\%.
The nonzero OOD performance indicates that the model can transfer some capabilities beyond the tasks providing supervision.
However, the substantial gap between ID and OOD performance shows that this transfer alone is insufficient for reliable execution across the OOD task set.

\textbf{Additional videos enable transfer without OOD action labels.}
Introducing OOD videos while retaining action supervision exclusively from ID tasks raises OOD success from 25.07\% to 55.90\%, an absolute improvement of 30.83 percentage points.
Meanwhile, ID performance remains nearly unchanged at 91.96\%.
The resulting overall success rate increases from 71.81\% to 81.14\%.
This improvement demonstrates that additional task knowledge supplied through videos can translate into better robot execution without corresponding OOD action labels.
It is consistent with the perspective illustrated in Figure~\ref{fig:task_generalization}: broader video coverage supports task transfer through a video-to-action bridge whose action supervision comes from ID tasks.
Nevertheless, the remaining gap between ID and OOD success suggests that video supervision alone does not fully resolve the challenges of executing these tasks.

\textbf{Sparse action supervision strengthens task-specific grounding.}
Keeping the video data fixed, adding only one action-labeled demonstration per OOD task increases OOD success from 55.90\% to 70.70\%.
With five demonstrations per task, success further rises to 84.67\%, yielding a 28.77-point improvement over the setting with OOD videos but no OOD action labels.
These gains are consistent with sparse paired supervision strengthening the inverse-dynamics capability that connects visual task evolution to executable actions.
In this regime, the additional demonstrations provide action grounding for tasks already covered by the video data.
Increasing the action budget to 20 demonstrations per task yields 90.03\% OOD success, only 2.80 points below the 550-demonstration setting while using approximately 3.6\% of its OOD action-label budget.
With 100 demonstrations, OOD success reaches 93.17\%; increasing the budget to 550 produces a comparable result of 92.83\%, suggesting diminishing returns at higher supervision levels.

\begin{table*}[t]
    \centering
    \caption{Average success rates (\%) on RoboTwin 2.0 under different amounts of OOD action supervision. OOD action counts denote the number of demonstrations per task.}
    \label{tab:ood_shot_results}
    \small
    \setlength{\tabcolsep}{7pt}
    \begin{tabular*}{\textwidth}{
        @{\extracolsep{\fill}}
        ccccccc
        @{\hspace{4em}}
        c@{}
    }
        \toprule
        \shortstack{\textbf{\strut Exp.}\\\textbf{\strut Index}}
        & \shortstack{\textbf{\strut \# ID}\\\textbf{\strut Videos}}
        & \shortstack{\textbf{\strut \# ID}\\\textbf{\strut Actions}}
        & \shortstack{\textbf{\strut \# OOD}\\\textbf{\strut Videos}}
        & \shortstack{\textbf{\strut \# OOD}\\\textbf{\strut Actions}}
        & \shortstack{\textbf{\strut ID}\\\textbf{\strut SR (\%)}}
        & \shortstack{\textbf{\strut OOD}\\\textbf{\strut SR (\%)}}
        & \shortstack{\textbf{\strut Overall}\\\textbf{\strut SR (\%)}} \\
        \midrule

        1 & 550 & 550 & 0 & 0
          & 91.85 & \makebox[2.5em][c]{25.07} & 71.81 \\

        2 & 550 & 550 & 550 & 0
          & 91.96 & \makebox[2.5em][c]{55.90}\makebox[0pt][l]{\hspace{0.5em}(+30.83)} & 81.14 \\

        3 & 550 & 550 & 550 & 1
          & 91.70 & \makebox[2.5em][c]{70.70}\makebox[0pt][l]{\hspace{0.5em}(+14.80)} & 85.40 \\

        4 & 550 & 550 & 550 & 5
          & 92.30 & \makebox[2.5em][c]{84.67}\makebox[0pt][l]{\hspace{0.5em}(+13.97)} & 90.01 \\

        5 & 550 & 550 & 550 & 20
          & 92.25 & \makebox[2.5em][c]{90.03}\makebox[0pt][l]{\hspace{0.5em}(+5.36)} & 91.58 \\

        6 & 550 & 550 & 550 & 50
          & 90.22 & \makebox[2.5em][c]{92.27}\makebox[0pt][l]{\hspace{0.5em}(+2.24)} & 90.83 \\

        7 & 550 & 550 & 550 & 100
          & 92.50 & \makebox[2.5em][c]{93.17}\makebox[0pt][l]{\hspace{0.5em}(+0.90)} & 92.70 \\

        8 & 550 & 550 & 100 & 100
          & 93.21 & \makebox[2.5em][c]{88.83} & 91.90 \\

        9 & 550 & 550 & 550 & 550
          & 91.55 & \makebox[2.5em][c]{92.83} & 91.93 \\

        \bottomrule
    \end{tabular*}
\end{table*}

\textbf{Scaling videos yields further gains with limited action supervision.}
Experiments 5--9 further illustrate the complementary roles of video and action supervision.
With 100 action-labeled demonstrations per OOD task, increasing OOD videos from 100 to 550 raises OOD success from 88.83\% to 93.17\%, a gain of 4.34 percentage points.
In contrast, with 550 OOD videos fixed, increasing the action budget from 100 to 550 demonstrations yields no improvement, from 93.17\% to 92.83\%.
Moreover, the setting with 550 videos and 20 action-labeled demonstrations (Exp. 5) outperforms that with 100 videos and 100 demonstrations (Exp.8), despite using one-fifth as many action labels.
These results suggest that a relatively small amount of paired supervision may suffice to learn an effective video-to-action bridge, or inverse dynamics model.
Once this action grounding is established, additional videos can further improve OOD task execution, whereas increasing action supervision offers limited additional gains in this regime.

\subsection{Simulation Benchmarks}

\begin{table*}[t]
    \centering
    \caption{Success rates (\%) on RoboTwin 2.0~\cite{pmlr-v306-chen26j}
    under two training-data settings. Each setting is evaluated in Clean and
    Randomized scenes, with their average reported. Clean-only training
    corresponds to the Clean2Random protocol.}
    \label{tab:robotwin2_results}
    \footnotesize
    \setlength{\tabcolsep}{5pt}
    \begin{tabular*}{\textwidth}{@{\extracolsep{\fill}}lccc@{\hspace{1.5em}}ccc}
        \toprule
        \multirow{2}{*}{\textbf{Method}}
        & \multicolumn{3}{c}{\textbf{Training: Clean + Randomized}}
        & \multicolumn{3}{c}{\textbf{Training: Clean Only}} \\
        \cmidrule(lr){2-4} \cmidrule(lr){5-7}
        & Clean & Randomized & Average
        & Clean & Randomized & Average \\
        \midrule

        GR00T-N1.7~\cite{nvidia2026grootn17}
            & -- & -- & -- & 43.6 & 20.7 & 32.2 \\

        StarVLA~\cite{community2026starvla}
            & -- & -- & -- & 58.1 & 10.6 & 34.4 \\

        Xiaomi-Robotics-0~\cite{cai2026xiaomi0}
            & -- & -- & -- & 62.9 & 18.2 & 40.6 \\

        $\pi_0$~\cite{black2024pi0}
            & 65.92 & 58.40 & 62.16 & -- & -- & -- \\

        X-VLA~\cite{zheng2026xvla}
            & 72.80 & 72.84 & 72.82 & 68.0 & 20.9 & 44.5 \\

        Spatial Forcing~\cite{spatialforcing2026}
            & -- & -- & -- & 77.2 & 26.7 & 52.0 \\

        $\pi_{0.5}$~\cite{physicalintelligence2025pi05}
            & 82.74 & 76.76 & 79.75 & 73.1 & 47.9 & 60.5 \\

        ABot-M0~\cite{yang2026abot}
            & 86.06 & 85.08 & 85.57 & 70.7 & 36.0 & 53.4 \\

        Qwen-VLA~\cite{wang2026qwenvla}
            & 86.10 & 87.20 & 86.65 & -- & -- & -- \\

        Motus~\cite{bi2026motus}
            & 88.66 & 87.02 & 87.84 & -- & -- & -- \\

        StarVLA-OFT~\cite{community2026starvla}
            & 88.20 & 88.30 & 88.25 & -- & -- & -- \\

        StarVLA-GR00T~\cite{community2026starvla}
            & 88.00 & 88.50 & 88.25 & -- & -- & -- \\

        StarVLA-$\pi$~\cite{community2026starvla}
            & 88.10 & 88.80 & 88.45 & -- & -- & -- \\

        JoyAI-RA 0.1~\cite{zhang2026joyaira}
            & 90.48 & 89.28 & 89.88 & -- & -- & -- \\

        Being-H0.7~\cite{luo2026beingh07}
            & 90.20 & 89.60 & 89.90 & -- & -- & -- \\

        X-WAM~\cite{guo2026xwam}
            & 89.80 & 90.70 & 90.25 & 70.0 & 25.8 & 47.9 \\

        Hy-Embodied-0.5-VLA~\cite{zhang2026hyembodied}
            & 90.90 & 90.10 & 90.50 & -- & -- & -- \\

        ACE-Ego-0~\cite{li2026aceego0}
            & 91.12 & 90.62 & 90.87 & -- & -- & -- \\

        4D-WAM~\cite{yang2026_4dwam}
            & 91.92 & 90.76 & 91.34 & 81.5 & 41.8 & 61.7 \\

        Fast-WAM~\cite{yuan2026fastwam}
            & 91.88 & 91.78 & 91.83 & 77.8 & 1.9 & 39.9 \\

        LingBot-VA~\cite{li2026causalworld}
            & 92.93 & 91.55 & 92.24 & 80.7 & 34.6 & 57.7 \\

        AHA-WAM~\cite{cai2026ahawam}
            & 93.40 & 92.20 & 92.80 & 64.3 & 3.2 & 33.8 \\

        InternVLA-A1.5~\cite{ma2026internvlaa15}
            & 93.30 & 93.00 & 93.20 & -- & -- & -- \\

        GalaxeaVLA (G0.5)~\cite{liu2026g05}
            & 93.70 & 92.80 & 93.30 & -- & -- & -- \\

        LingBot-VA 2.0~\cite{zhang2026native}
            & 93.80 & 93.40 & 93.60 & -- & -- & -- \\

        GigaBrain-0.7~\cite{gigabrain2026}
            & -- & -- & -- & 66.8 & 67.9 & 67.4 \\

        OpenWAM-$\alpha$~\cite{wang2026openwam}
            & 93.74 & 93.46 & 93.60 & 89.4 & 48.7 & 69.0 \\

        Qwen-RobotManip~\cite{qwenrobotmanip2026}
            & 93.70 & 94.00 & 93.85 & -- & -- & -- \\

        Long-WAM (IDM)~\cite{huang2026longwam}
            & 94.70 & 94.20 & 94.40 & -- & -- & -- \\

        MotuBrain~\cite{motubrainteam2026motubrain}
            & 95.80 & 96.08 & 95.94 & -- & -- & -- \\

        Kairos~\cite{kairosteam2026kairos}
            & 96.90 & 95.20 & 96.05 & -- & -- & -- \\
        
        Qwen-RobotManip-Context~\cite{qwenrobotmanip2026}
            & -- & -- & -- & 84.7 & 69.4 & 77.1 \\

        \midrule
        \textbf{WorldBridge (Ours)}
            & 96.16 & 96.30 & \textbf{96.23}
            & 88.3 & 73.4 & \textbf{80.9} \\
        \bottomrule
    \end{tabular*}
\end{table*}

% Requires \usepackage{booktabs,multirow,graphicx}
\begin{table*}[t]
    \centering
    \caption{Results on RoboCasa GR1 TableTop, RMBench, DOMINO Level-1,
    and SimplerEnv-WidowX. We report success rates (\%) and the
    DOMINO manipulation score (MS)}
    \label{tab:simulation_benchmarks}
    \label{tab:robocasa_gr1_results}
    \label{tab:simplerenv_results}
    \label{tab:rmbench_results}
    \label{tab:domino_results}
    \footnotesize
    \setlength{\tabcolsep}{3.5pt}
    \resizebox{\textwidth}{!}{%
    \begin{tabular}{lccccccccc}
        \toprule
        \multirow{2}{*}{\textbf{Method}}
        & \multicolumn{1}{c}{\textbf{RoboCasa GR1}}
        & \multicolumn{1}{c}{\textbf{RMBench}}
        & \multicolumn{2}{c}{\textbf{DOMINO}}
        & \multicolumn{5}{c}{\textbf{SimplerEnv-WidowX}} \\
        \cmidrule(lr){2-2} \cmidrule(lr){3-3}
        \cmidrule(lr){4-5} \cmidrule(lr){6-10}
        & SR & SR & SR & MS
        & Spoon & Carrot & Cube & Eggplant & Avg. \\
        \midrule

        \multicolumn{10}{c}{\textit{VLA}} \\
        \midrule

        OpenVLA~\cite{kim2024openvla}
            & -- & -- & 1.5 & 6.1
            & 4.2 & 0.0 & 0.0 & 12.5 & 4.2 \\

        X-VLA~\cite{zheng2026xvla,chen2026rmbench}
            & -- & 9.8 & -- & --
            & 100.0 & 91.7 & 95.8 & 95.8 & 95.8 \\

        $\pi_0$~\cite{black2024pi0}
            & 13.6 & -- & 8.2 & 24.0
            & 29.1 & 0.0 & 16.6 & 62.5 & 27.1 \\

        PUMA~\cite{fang2026domino}
            & -- & -- & 17.2 & 35.0
            & -- & -- & -- & -- & -- \\

        Magma~\cite{yang2025magma}
            & -- & -- & -- & --
            & 37.5 & 31.0 & 12.7 & 60.5 & 35.8 \\

        $\pi_{0.5}$~\cite{physicalintelligence2025pi05}
            & 37.0 & 10.4 & 9.6 & 26.2
            & 49.3 & 64.7 & 44.7 & 69.7 & 57.1 \\

        Mem-0~\cite{chen2026rmbench}
            & -- & 42.0 & -- & --
            & -- & -- & -- & -- & -- \\

        SpatialVLA~\cite{qu2025spatialvla}
            & -- & -- & -- & --
            & 16.7 & 25.0 & 29.2 & 100.0 & 42.7 \\

        CronusVLA~\cite{li2025cronusvla}
            & -- & -- & -- & --
            & 66.7 & 54.2 & 20.8 & 100.0 & 60.4 \\

        GR00T-N1.6~\cite{nvidia2025groot}
            & 47.6 & -- & -- & --
            & 55.4 & 46.0 & 5.0 & 89.0 & 48.9 \\

        StarVLA-$\pi$~\cite{community2026starvla}
            & 48.9 & -- & -- & --
            & 78.1 & 46.9 & 30.2 & 88.5 & 60.9 \\

        RoboInter~\cite{li2026robointer}
            & -- & -- & -- & --
            & 54.2 & 25.0 & 29.2 & 91.7 & 50.0 \\

        StarVLA-GR00T~\cite{community2026starvla}
            & 52.8 & -- & -- & --
            & 83.0 & 59.4 & 18.8 & 100.0 & 65.3 \\

        StarVLA-OFT~\cite{community2026starvla}
            & 53.8 & -- & 10.9 & 30.5
            & 90.3 & 38.5 & 29.7 & 100.0 & 64.6 \\

        Qwen-VLA-Instruct~\cite{wang2026qwenvla}
            & 56.7 & -- & 26.6 & 39.5
            & -- & -- & -- & -- & 73.7 \\

        ABot-M0~\cite{yang2026abot}
            & 58.3 & -- & -- & --
            & -- & -- & -- & -- & -- \\

        GR00T-N1.7~\cite{nvidia2026grootn17}
            & -- & -- & -- & --
            & 78.0 & 58.0 & 48.0 & 53.0 & 59.3 \\

        JoyAI-RA 0.1~\cite{zhang2026joyaira}
            & 63.2 & -- & -- & --
            & -- & -- & -- & -- & -- \\

        DIAL~\cite{chen2026dial}
            & 70.2 & -- & -- & --
            & -- & -- & -- & -- & -- \\

        FutureVLA~\cite{xu2026futurevla}
            & -- & -- & -- & --
            & 83.3 & 75.0 & 29.2 & 100.0 & 71.9 \\

        ACE-Ego-0~\cite{li2026aceego0}
            & 72.8 & -- & -- & --
            & -- & -- & -- & -- & -- \\

        LaMem-VLA~\cite{qu2026lamemvla}
            & -- & -- & -- & --
            & 83.3 & 75.0 & 41.7 & 95.8 & 73.9 \\

        InternVLA-A1.5~\cite{ma2026internvlaa15}
            & -- & -- & 29.3 & 42.5
            & 92.4 & 85.4 & 69.5 & 75.7 & 80.8 \\
            
        Xiaomi-Robotics-0~\cite{cai2026xiaomi0}
            & -- & -- & -- & --
            & 95.8 & 62.5 & 75.0 & 83.3 & 79.2 \\

        GTA-VLA~\cite{ling2026gtavla}
            & -- & -- & -- & --
            & 95.8 & 87.5 & 66.7 & 75.0 & 81.2 \\

        GalaxeaVLA (G0.5)~\cite{liu2026g05}
            & -- & -- & -- & --
            & 97.5 & 75.0 & 83.3 & 93.3 & 87.3 \\

        \midrule
        \multicolumn{10}{c}{\textit{WAM}} \\
        \midrule

        Fast-WAM~\cite{yuan2026fastwam}
            & -- & 5.9 & 19.9 & 33.3
            & -- & -- & -- & -- & -- \\
        OA-WAM~\cite{liu2026oa}
            & -- & -- & -- & --
            & 83.0 & 71.1 & 65.0 & 98.2 & 79.3 \\

        DynamicWAM~\cite{lou2026dynamicwam}
            & -- & -- & 38.2 & 53.2
            & -- & -- & -- & -- & -- \\

        Being-H0.7~\cite{luo2026beingh07}
            & 49.2 & -- & -- & --
            & -- & -- & -- & -- & -- \\

        OpenWAM-$\alpha$~\cite{wang2026openwam}
            & 60.5 & -- & -- & --
            & -- & -- & -- & -- & -- \\

        LingBot-VA~\cite{li2026causalworld}
            & -- & 78.2 & 24.1 & 36.1
            & -- & -- & -- & -- & -- \\

        Long-WAM~\cite{huang2026longwam}
            & 78.7 & -- & 34.9 & 45.1
            & -- & -- & -- & -- & -- \\

        MemoryWAM~\cite{yang2026memorywam}
            & -- & \textbf{83.0} & -- & --
            & -- & -- & -- & -- & -- \\

        \midrule
        \textbf{WorldBridge (Ours)}
            & \textbf{82.8} & \textbf{83.0}
            & \textbf{57.6} & \textbf{64.6}
            & 95.8 & 83.3 & 91.7 & 66.7 & \textbf{84.4} \\
        \bottomrule
    \end{tabular}%
    }
\end{table*}

\begin{table*}[t]
\centering
\caption{Performance comparison on the RoboDojo benchmark.
Score and success rate (SR, \%) are reported across
six evaluation settings and averaged over the five
capability dimensions. Additional entries are taken from the official leaderboard (October 9, 2026)~\cite{robodojo2026leaderboard}.}
\label{tab:robodojo_results}
\scriptsize
\setlength{\tabcolsep}{3.5pt}
\resizebox{\textwidth}{!}{
\begin{tabular}{l*{7}{cc}}
\toprule
\multirow{2}{*}{\textbf{Model}}
& \multicolumn{2}{c}{\textbf{Gen-Std}}
& \multicolumn{2}{c}{\textbf{Gen-Rand}}
& \multicolumn{2}{c}{\textbf{Precision}}
& \multicolumn{2}{c}{\textbf{Long}}
& \multicolumn{2}{c}{\textbf{Memory}}
& \multicolumn{2}{c}{\textbf{Open}}
& \multicolumn{2}{c}{\textbf{Average}} \\
\cmidrule(lr){2-3}
\cmidrule(lr){4-5}
\cmidrule(lr){6-7}
\cmidrule(lr){8-9}
\cmidrule(lr){10-11}
\cmidrule(lr){12-13}
\cmidrule(lr){14-15}
& Score & SR & Score & SR
& Score & SR & Score & SR
& Score & SR & Score & SR
& Score & SR \\
\midrule

\multicolumn{15}{c}{\textit{VLA}} \\
\midrule

        % ADDED: https://robodojo-benchmark.com/data/leaderboard.generated.json (updated 2026-10-09)
        GR00T-N1.7~\cite{nvidia2026grootn17}
            & 3.97 & 2.44 & 0.35 & 0.00 & 2.54 & 0.67 & 8.30 & 3.58 & 1.06 & 0.89 & 0.18 & 0.17 & 2.85 & 1.31 \\
        % ADDED: https://robodojo-benchmark.com/data/leaderboard.generated.json (updated 2026-10-09)
        $\pi_0$~\cite{black2024pi0}
            & 7.18 & 4.89 & 0.71 & 0.22 & 3.56 & 0.75 & 6.19 & 2.00 & 3.47 & 2.11 & 0.25 & 0.25 & 3.48 & 1.53 \\
        % ADDED: https://robodojo-benchmark.com/data/leaderboard.generated.json (updated 2026-10-09)
        ABot-M0~\cite{yang2026abot}
            & 9.20 & 5.44 & 2.26 & 1.56 & 5.50 & 1.75 & 3.96 & 0.50 & 2.44 & 2.22 & 0.72 & 0.67 & 3.67 & 1.73 \\
        % ADDED: https://robodojo-benchmark.com/data/leaderboard.generated.json (updated 2026-10-09)
        StarVLA~\cite{community2026starvla}
            & 7.54 & 4.67 & 0.33 & 0.00 & 9.90 & 4.33 & 14.15 & 6.50 & 3.34 & 2.44 & 0.67 & 0.58 & 6.40 & 3.24 \\
        % ADDED: https://robodojo-benchmark.com/data/leaderboard.generated.json (updated 2026-10-09)
        Xiaomi-Robotics-0~\cite{cai2026xiaomi0}
            & 13.81 & 10.78 & 1.05 & 0.33 & 8.42 & 4.58 & 13.51 & 6.92 & 5.07 & 3.67 & 0.22 & 0.17 & 6.93 & 4.18 \\
        % ADDED: https://robodojo-benchmark.com/data/leaderboard.generated.json (updated 2026-10-09)
        StarVLA-OFT~\cite{community2026starvla}
            & 13.42 & 9.00 & 1.39 & 0.56 & 16.08 & 10.17 & 14.28 & 6.17 & 1.74 & 1.67 & 1.41 & 1.33 & 8.18 & 4.82 \\
        % ADDED: https://robodojo-benchmark.com/data/leaderboard.generated.json (updated 2026-10-09)
        StarVLA-GR00T~\cite{community2026starvla}
            & 13.94 & 9.56 & 1.97 & 1.00 & 9.56 & 4.67 & 16.18 & 7.17 & 5.60 & 4.33 & 1.95 & 1.92 & 8.25 & 4.67 \\
        % ADDED: https://robodojo-benchmark.com/data/leaderboard.generated.json (updated 2026-10-09)
        X-VLA~\cite{zheng2026xvla}
            & 17.90 & 12.22 & 3.04 & 1.33 & 18.32 & 12.00 & 16.53 & 9.75 & 4.76 & 3.56 & 0.55 & 0.50 & 10.13 & 6.52 \\

StarVLA-PI\_v3~\cite{community2026starvla}
& 19.08 & 14.44 & 3.36 & 1.67
& 17.77 & 12.50 & 18.46 & 11.00
& 4.59 & 4.00 & 2.03 & 2.00
& 10.81 & 7.51 \\

InternVLA-A1.5~\cite{ma2026internvlaa15}
& 16.81 & 11.78 & 3.90 & 1.89
& 15.23 & 10.17 & 23.80 & 13.75
& 4.93 & 3.56 & 1.43 & 1.42
& 11.15 & 7.14 \\

$\pi_{0.5}$~\cite{physicalintelligence2025pi05}
& 20.93 & 14.89 & 5.82 & 1.44
& 12.40 & 5.50 & 23.54 & 14.67
& 5.78 & 4.56 & 1.98 & 1.67
& 11.41 & 6.91 \\

Spatial Forcing~\cite{spatialforcing2026}
& 21.25 & 14.89 & 6.98 & 3.78
& 17.32 & 10.58 & 23.26 & 14.58
& 5.43 & 4.11 & 1.78 & 1.58
& 12.38 & 8.04 \\

Hy-Embodied-0.5-VLA~\cite{zhang2026hyembodied}
& 21.98 & 16.56 & 1.57 & 0.22
& 13.81 & 8.00 & 25.74 & 14.92
& 13.37 & 12.11 & 0.65 & 0.58
& 13.07 & 8.80 \\

Xiaomi-Robotics-1~\cite{team2026xiaomi}
& 35.65 & 28.00 & 11.44 & 6.00
& 26.69 & 18.83 & 38.39 & 23.67
& 7.81 & 6.56 & 3.94 & 3.58
& 20.07 & 13.93 \\

GalaxeaVLA (G0.5)~\cite{liu2026g05}
& 27.72 & 21.44 & 9.20 & 4.22
& 28.25 & 20.42 & 44.12 & 32.25
& 8.61 & 7.33 & 1.73 & 1.58
& 20.23 & 14.88 \\

DM0.5~\cite{dm052026}
& 23.49 & 17.89 & 8.06 & 4.00
& 24.82 & 16.75 & 33.70 & 19.50
& 47.74 & 47.44 & 2.43 & 2.08
& 24.90 & 19.34 \\

\midrule
\multicolumn{15}{c}{\textit{WAM}} \\
\midrule

        % ADDED: https://robodojo-benchmark.com/data/leaderboard.generated.json (updated 2026-10-09)
        Fast-WAM~\cite{yuan2026fastwam}
            & 4.33 & 2.11 & 0.34 & 0.11 & 1.96 & 0.00 & 9.14 & 5.17 & 3.55 & 3.44 & 0.42 & 0.42 & 3.48 & 2.03 \\
        % ADDED: https://robodojo-benchmark.com/data/leaderboard.generated.json (updated 2026-10-09)
        AHA-WAM~\cite{cai2026ahawam}
            & 10.32 & 6.22 & 1.26 & 0.33 & 5.86 & 2.42 & 8.61 & 2.67 & 2.97 & 2.78 & 0.88 & 0.83 & 4.82 & 2.39 \\
        % ADDED: https://robodojo-benchmark.com/data/leaderboard.generated.json (updated 2026-10-09)
        X-WAM~\cite{guo2026xwam}
            & 11.24 & 5.33 & 3.54 & 1.33 & 6.72 & 1.83 & 17.47 & 9.08 & 6.32 & 4.67 & 0.57 & 0.25 & 7.69 & 3.83 \\

OpenWAM-$\alpha$~\cite{wang2026openwam}
& 33.16 & 25.56 & 8.26 & 4.11
& 18.45 & 9.25 & 34.93 & 25.33
& 10.41 & 9.11 & 1.41 & 1.08
& 17.18 & 11.92 \\

\textbf{WorldBridge (Ours)}
& 37.75 & 30.00 & 11.35 & 4.00 & 35.40 & 25.25 & 42.25 & 27.00 & 32.93 & 32.67 & 4.60 & 3.75 & \textbf{27.95} & \textbf{21.13} \\ 
\bottomrule
\end{tabular}
}
\end{table*}

\textbf{RoboTwin 2.0.} We evaluate on 50 language-conditioned manipulation tasks using the ALOHA-AgileX bimanual robot in RoboTwin 2.0~\cite{pmlr-v306-chen26j}, covering object rearrangement, articulated-object manipulation, and coordinated two-arm control. We consider two settings: Full RoboTwin uses 50 Clean and 500 Randomized demonstrations per task, while Clean2Random uses only the 50 Clean demonstrations to assess generalization to randomized scenes of the same tasks. The policy conditions on head and wrist RGB observations, language instructions, and proprioception, and incorporates short- and long-term visual memory to predict joint-position and gripper commands. We train for 60,000 steps on the combined Clean and Randomized data in Full RoboTwin, and for 20,000 steps on Clean data alone in Clean2Random. We evaluate each setting over 100 episodes per task per scene condition, totaling 10,000 trials. As shown in Table~\ref{tab:robotwin2_results}, WorldBridge achieves 96.16\% success on Clean and 96.30\% on Randomized scenes under Full RoboTwin, averaging 96.23\%. Under Clean2Random, it achieves 88.3\% on Clean and 73.4\% on Randomized scenes, averaging 80.9\%. These results demonstrate reliable execution across diverse manipulation tasks and robustness to scene variations, including transfer to randomized environments when downstream training uses only Clean demonstrations.

\textbf{RoboCasa GR1.}
We train on 24,000 teleoperated demonstrations spanning 24 language-conditioned manipulation tasks with the Fourier GR1 humanoid robot in RoboCasa~\cite{nasiriany2024robocasa}, with 1,000 episodes per task. These tasks cover pick-and-place and articulated-object manipulation. We find that video generation quality is important for downstream policy performance. We therefore replay the original demonstrations at a higher rendering resolution, crop out the black borders at the top and bottom, and resize the resulting observations to $416 \times 288$ pixels (width $\times$ height). We fine-tune the model for 100K steps on 16 GPUs, using learning rates of $5\times10^{-5}$ for the video backbone and $1\times10^{-5}$ for the action expert. For evaluation, we perform 50 rollouts on each of the 24 tasks, totaling 1,200 trials, and achieve an overall success rate of 82.75\%, outperforming the strongest VLA~\cite{li2026aceego0} and WAM~\cite{huang2026longwam} baselines in our comparison by 10.0 and 4.1 percentage points, respectively.

\textbf{RoboDojo.}
We evaluate our policy on RoboDojo~\cite{chen2026robodojo}, an Isaac Sim-based
benchmark for bimanual manipulation using the ARX X5 platform.
The benchmark comprises 42 tasks spanning five complementary
capability dimensions: Generalization, Precision, Long-Horizon,
Memory, and Open. These dimensions assess robustness to scene
variations, fine-grained manipulation accuracy, multi-step execution,
reasoning over past observations, and transfer to unseen task
specifications, respectively. Our policy achieves an overall Score of 27.95 and a success rate
of 21.13\% on RoboDojo, with Scores of 24.55, 35.40, 42.25, 32.93,
and 4.60 across the five capability dimensions: Generalization,
Precision, Long-Horizon, Memory, and Open, respectively.
These results demonstrate capabilities spanning fine-grained
control, multi-stage manipulation, and reasoning over observation
history. We fine-tune the policy using AdamW with a learning rate of
$5\times10^{-5}$, a global batch size of 256, and bfloat16 precision.
We use a cosine learning-rate schedule with 2,000 warm-up steps
and report results from the checkpoint after 60,000 fine-tuning steps.

\textbf{SimplerEnv-WidowX.} We evaluate language-conditioned manipulation policy transfer on the WidowX suite of SimplerEnv~\cite{li2024evaluating}, a real-to-sim benchmark that instantiates the BridgeData V2/WidowX setup for scalable simulated policy evaluation. We follow the official SimplerEnv WidowX task suite and evaluate four Bridge-style manipulation tasks: putting a spoon on a towel, putting a carrot on a plate, stacking a green cube on a yellow cube, and putting an eggplant into a yellow basket. WorldBridge is fine-tuned on the real-world dataset BridgeData V2~\cite{walke2023bridgedata} for 60K gradient steps, without additional simulation demonstrations during this adaptation. The policy conditions on RGB observations, language instructions, and proprioception, with short- and long-term visual histories, and predicts 16 consecutive 7-D actions. As shown in Table~\ref{tab:simplerenv_results}, WorldBridge achieves 84.4\% average success, with task-wise rates of 95.8\%, 83.3\%, 91.7\%, and 66.7\%, respectively. It exceeds the OA-WAM, InternVLA-A1.5 and the most VLAs with a reported result in this suite.

\textbf{RMBench.} RMBench~\cite{chen2026rmbench} assesses dual-arm manipulation across nine non-Markovian tasks where decision-making relies heavily on past temporal cues rather than instantaneous observations. 
Following the standard protocol, we train the policy with $50$ expert trajectories per task and evaluate over $100$ rollouts. 
Crucially, during both training and inference, we cache all historical frames to serve as the conditioning context. 
As reported in Table~\ref{tab:rmbench_results}, methods limited by bounded observation windows or reactive mappings struggle severely under such temporal dependencies, with Fast-WAM~\cite{yuan2026fastwam} ($5.9\%$), X-VLA~\cite{zheng2026xvla} ($9.8\%$), and $\pi_{0.5}$~\cite{physicalintelligence2025pi05} ($10.4\%$) achieving marginal success. 
In contrast, {WorldBridge} establishes state-of-the-art performance with an \textbf{83.0\%} average success rate---surpassing strong full-history baselines such as LingBot-VA~\cite{li2026causalworld} ($78.2\%$) and matching MemoryWAM~\cite{yang2026memorywam}---demonstrating its capability to leverage complete temporal contexts for complex robotic manipulation.

\textbf{DOMINO.} This benchmark~\cite{fang2026domino} evaluates manipulation with moving targets, extending static-scene evaluation to dynamic interception and tracking. We evaluate the ALOHA-AgileX on all 35 Level-1 tasks, which use constant-velocity target motion with dynamics coefficient $\alpha=0.1$, and evaluate 7,000 episodes in total across the Clean and Randomized dynamic configurations, following the PUMA evaluation protocol~\cite{fang2026domino}. WorldBridge directly reuses the checkpoint trained on RoboTwin 2.0 and is evaluated on DOMINO without additional dynamic-task fine-tuning. The policy uses head and wrist RGB observations, language, and proprioception; it predicts 16 actions and executes the first ten before replanning. Randomized evaluation additionally varies backgrounds, lighting, table height, and clutter. We report success rate (SR) and manipulation score (MS), averaged over both configurations, and MS measures normalized end-effector progress toward the target, with penalties for out-of-bounds motion and clutter collisions. Table~\ref{tab:simplerenv_results} reports 57.6\% SR and 64.6 MS for WorldBridge, exceeding the reported results of DynamicWAM, the second-best method in this comparison, by 19.4 percentage points in SR and 11.4 points in MS. Within our evaluation, Clean yields 59.1\% SR and 67.7 MS, while Randomized yields 56.1\% SR and 61.6 MS, showing a modest success-rate decrease under additional scene variation. These results demonstrate strong static-to-dynamic transfer of our model without dynamic-task fine-tuning. This capability benefits from the complementary roles of short- and long-term memory, which provide temporal cues and world-modeling priors acquired through pretraining.

\textbf{Benefits of action forcing and memory.}
Table~\ref{tab:history_ablation} presents a cumulative ablation, adding action forcing, short-term memory, and long-term memory in sequence. Following the joint video--action architecture in Section~\ref{sec:architecture}, action forcing trains the action expert with actions at different noise levels within a chunk (Section~\ref{sec:action_forcing}), while the two memory scales provide recent motion cues and broader task history (Section~\ref{sec:memory}). Adding action forcing alone improves success from 91.28\% to 93.31\% on RoboTwin, from 67.50\% to 69.17\% on RoboCasa, and from 14.50\% to 16.52\% on RoboDojo. Adding short-term memory next yields further gains of 0.67, 5.16, and 0.38 percentage points, respectively; adding long-term memory then contributes another 1.22, 4.25, and 4.17 points. The full configuration reaches 95.20\%, 78.58\%, and 21.07\%, improving over the configuration without action forcing or memory by 3.92, 11.08, and 6.57 points. These sequential results support the complementary value of action forcing and temporal context, with memory additions particularly beneficial on RoboCasa and RoboDojo.

\begin{table*}[t!]
    \centering
    \caption{Ablation study of action forcing and memory length on three robotic manipulation benchmarks.}
    \label{tab:history_ablation}
    \small
    \setlength{\tabcolsep}{7pt}

    \begin{tabular*}{\textwidth}{
        @{\extracolsep{\fill}}
        cccc
        @{\hspace{2em}}
        ccc
        @{}
    }
        \toprule
        \shortstack{\textbf{Exp.}\\\textbf{Index}}
        & \shortstack{\textbf{Action}\\\textbf{Forcing}}
        & \shortstack{\textbf{Short}\\\textbf{Memory}}
        & \shortstack{\textbf{Long}\\\textbf{Memory}}
        & \shortstack{\textbf{RoboTwin}\\\textbf{SR (\%)}}
        & \shortstack{\textbf{RoboCasa}\\\textbf{SR (\%)}}
        & \shortstack{\textbf{RoboDojo}\\\textbf{Score (\%)}} \\
        \midrule

        1 & \xmark & \xmark & \xmark
          & 91.28 & 67.50 & 14.50 \\

        2 & \cmark & \xmark & \xmark
          & 93.31 & 69.17 & 16.52 \\

        3 & \cmark & \cmark & \xmark
          & 93.98 & 74.33 & 16.90 \\

        4 & \cmark & \cmark & \cmark
          & {95.20} & {78.58} & {21.07} \\

        \bottomrule
    \end{tabular*}
\end{table*}

% \begin{table*}[t]
%     \centering
%     \caption{Effect of pretraining on downstream robotic manipulation benchmarks.}
%     \label{tab:pretrain_ablation}
%     \small
%     \setlength{\tabcolsep}{7pt}

%     \begin{tabular*}{\textwidth}{
%         @{\extracolsep{\fill}}
%         lcccccc@{}
%     }
%         \toprule
%         \textbf{Configuration}
%         & \shortstack{\textbf{RoboTwin}\\\textbf{SR (\%)}}
%         & \shortstack{\textbf{RoboCasa}\\\textbf{SR (\%)}}
%         & \shortstack{\textbf{RoboDojo}\\\textbf{SR (\%)}}
%         & \shortstack{\textbf{RMBench}\\\textbf{SR (\%)}}
%         & \shortstack{\textbf{Simpler}\\\textbf{SR (\%)}}
%         & \shortstack{\textbf{Domino}\\\textbf{SR (\%)}} \\
%         \midrule

%         w/o pretrain
%         & 91.76 & 78.58 & 21.07 & 81.00 & -- & -- \\

%         w/ pretrain
%         & {95.22}
%         & {82.83}
%         & {26.03}
%         & {83.00}
%         & --
%         & -- \\

%         \bottomrule
%     \end{tabular*}
% \end{table*}

% 导言区添加：\usepackage{booktabs}

\begin{table*}[t]
    \centering
    \caption{Effect of robot and egocentric pretraining on RoboTwin Clean-to-Random.}
    \label{tab:pretrain_clean2rand}
    \renewcommand{\arraystretch}{1.15}
    \begin{tabular*}{\textwidth}{@{\extracolsep{\fill}}cccccc@{}}
        \toprule
        \multirow{2}{*}{\textbf{Exp. Index}}
        & \multicolumn{2}{c}{\textbf{Pretraining}}
        & \multicolumn{3}{c}{\textbf{RoboTwin Clean-to-Random}} \\
        \cmidrule(lr){2-3}
        \cmidrule(l){4-6}
        & Robot
        & Egocentric
        & Clean (\%)
        & Rand (\%)
        & Avg (\%) \\
        \midrule
        1 & \xmark & \xmark & 82.9 & 42.6 & 62.7 \\
        2 & \cmark & \xmark & 87.0 & 67.7 & 77.4 \\
        3 & \cmark & \cmark
          & {88.3} & {73.4} & {80.9} \\
        \bottomrule
    \end{tabular*}
\end{table*}

\textbf{Benefits of pretraining.}
Table~\ref{tab:pretrain_clean2rand} isolates the contributions of robot and egocentric pretraining on RoboTwin Clean-to-Random. Without either source, the model achieves 62.7\% average success (82.9\% on Clean and 42.6\% on Randomized scenes). Adding robot pretraining raises the average to 77.4\% (+14.7 percentage points), with the largest gain on Randomized scenes, from 42.6\% to 67.7\%. Adding egocentric pretraining on top further improves the Clean and Randomized results to 88.3\% and 73.4\%, respectively, raising the average to 80.9\% (+3.5 points). Overall, combining both sources improves average success by 18.2 points over training without pretraining, suggesting that egocentric data provides complementary gains, particularly under scene randomization.

\subsection{Real-Robot Experiments}
\label{sec:real_robot_experiments}

\begin{figure}[t]
    \centering
    \includegraphics[width=\textwidth]{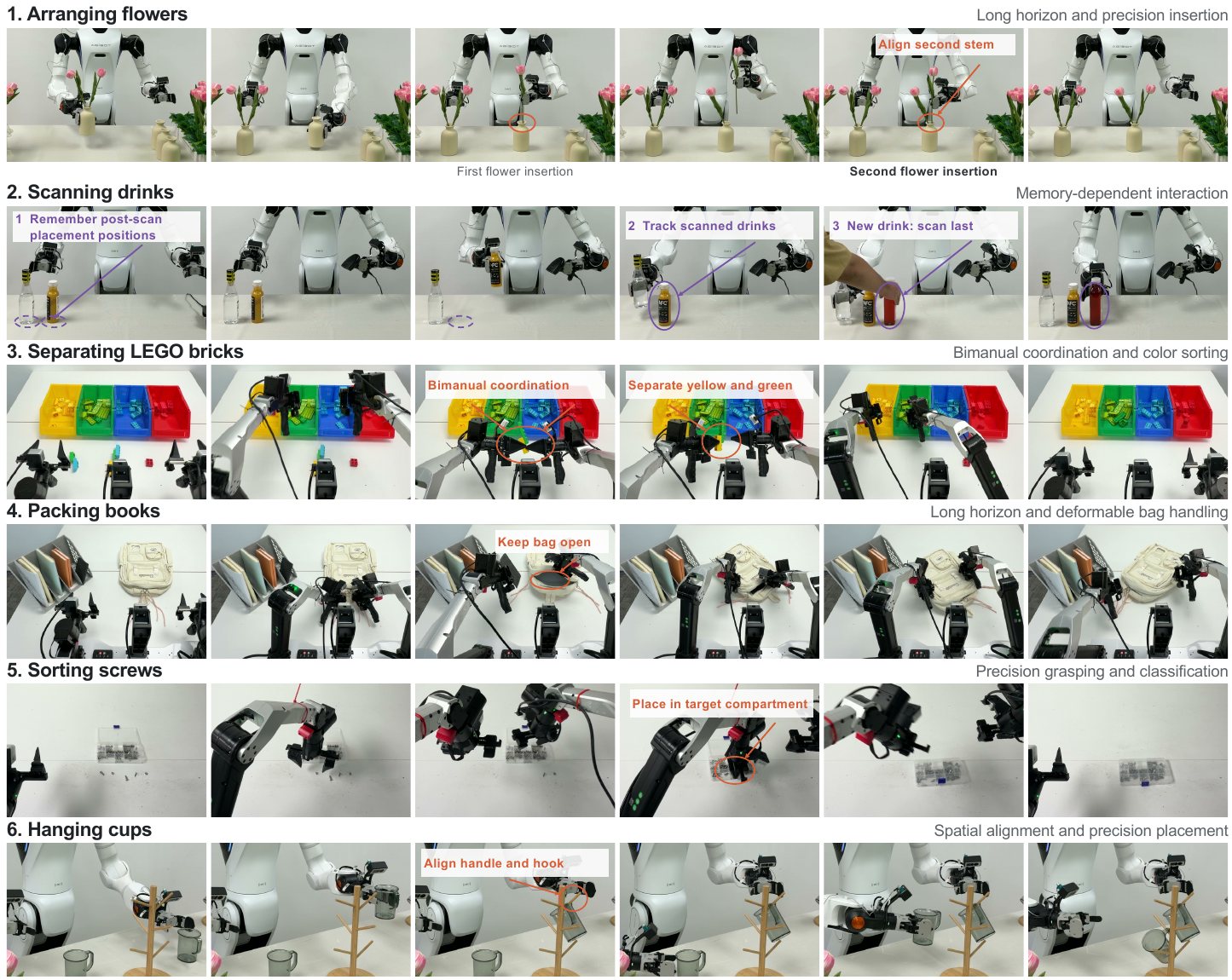}
    \caption{Representative real-robot executions on six manipulation tasks. From top to bottom: arranging flowers, scanning drinks, separating LEGO bricks, packing books, sorting screws, and hanging cups. Arrows and circles highlight key operations, including sequential flower insertion, bimanual brick separation, and precise placement. The drink-scanning task probes three aspects of memory: remembering post-scan placement destinations, tracking previously scanned drinks, and processing the newly introduced drink last.}
    \label{fig:real_robot_task_execution}
\end{figure}

\begin{figure}[t]
    \centering
    \includegraphics[width=\textwidth]{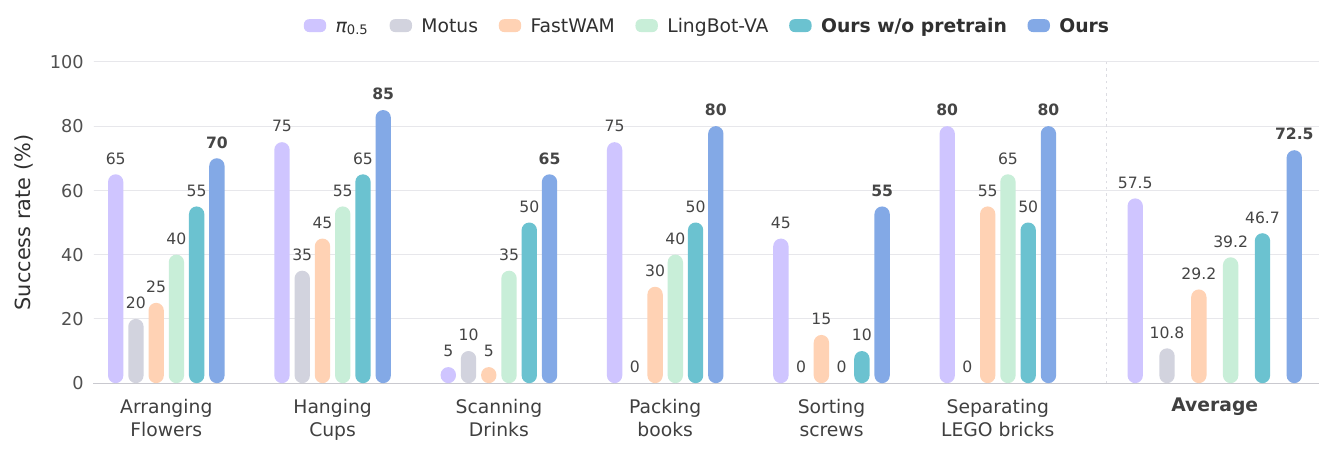}
    \caption{Task success rates across five methods.}
    \label{fig:real_robot_task_success_rates}
\end{figure}

\paragraph{Tasks and evaluation procedure.}
We evaluate WorldBridge on six complementary manipulation tasks across AgiBot G2 and ARX AC-1, covering long-horizon execution, precise manipulation, spatial alignment, bimanual coordination, and memory-dependent interaction.
Figure~\ref{fig:real_robot_task_execution} presents representative executions, with annotations highlighting the operations associated with these capability requirements.
On G2, \emph{arranging flowers} requires relocating vases, successively grasping and inserting two slender flowers, and placing the completed arrangement.
It tests long-horizon sequencing together with precision insertion, since the robot must preserve task progress while repeatedly aligning a thin stem with a vase opening.
\emph{Scanning drinks} interleaves sequential manipulation with human-induced scene changes, testing whether the robot can maintain task continuity and the required processing order using visual memory.
\emph{Hanging cups} requires grasping two mugs and placing their handles onto rack hooks at different heights.
It tests spatial understanding and orientation-sensitive placement, coupling grasp selection with alignment of the handle and hook.

On AC-1, \emph{separating LEGO bricks} requires repeated bimanual disassembly of connected bricks followed by sorting into bins by color.
It tests long-horizon coordination, stable grasping during separation, and classification-guided placement as the assembly changes throughout execution.
\emph{Sorting screws} requires grasping small screws and nuts, placing them into their designated compartments, and closing the box.
It combines fine-grained object discrimination with precise grasping and placement, requiring correct part selection and accurate manipulation within a single sequence.
\emph{Packing books} requires opening a schoolbag, inserting a book while maintaining access to the opening, and final closing the bag.
It tests long-horizon manipulation of a deformable container: the robot must coordinate bag opening with subsequent insertion despite changes in the bag's shape and accessible geometry.

We compare WorldBridge with $\pi_{0.5}$~\cite{physicalintelligence2025pi05}, Motus~\cite{bi2026motus}, FastWAM~\cite{yuan2026fastwam}, and LingBot-VA~\cite{li2026causalworld}, and evaluate a WorldBridge variant without our joint video--action pretraining.
All methods are fine-tuned on the same demonstration set for each task and evaluated over 20 trials with varied initial object positions or orientations.
Success requires completing the entire task without human assistance to the robot; introducing a drink during scanning is a prescribed interaction.
We report per-task and average success rates across all tasks.

\paragraph{Overall real-robot performance.}
Figure~\ref{fig:real_robot_task_success_rates} summarizes the quantitative results.
WorldBridge achieves an average success rate of 72.5\%, compared with 57.5\% for $\pi_{0.5}$, 39.2\% for LingBot-VA, 29.2\% for FastWAM, and 10.8\% for Motus.
It achieves the highest success rate on five tasks and ties $\pi_{0.5}$ at 80\% on separating LEGO bricks.
On G2, WorldBridge reaches 70\% on arranging flowers and 85\% on hanging cups, compared with 65\% and 75\% for $\pi_{0.5}$, respectively.
These results cover both extended manipulation sequences and precise spatial placement; the larger gap on scanning drinks is discussed below.
On AC-1, WorldBridge achieves 80\% on packing books and 55\% on sorting screws, exceeding $\pi_{0.5}$ at 75\% and 45\%, respectively, while matching its LEGO disassembly performance.
Sorting screws remains the least successful task for WorldBridge, identifying precise small-part sorting as a remaining challenge within this suite.

\paragraph{Cross-embodiment experiments.}
Cross-embodiment adaptation follows our unified 46-dimensional action representation, which maps robot-specific control variables into fixed semantic slots and uses dimension-wise validity masks to exclude unavailable components from pretraining supervision.
This common interface allows the action expert to learn from heterogeneous control data while preserving the semantics of the available controls.
For the real-robot experiments, we fine-tune the same jointly pretrained checkpoint separately for G2 and AC-1, transferring the shared video and action experts and adapting the state and action interfaces to each platform.
G2 uses 16-dimensional native commands comprising seven joints per arm and two grippers, whereas AC-1 uses 14-dimensional commands comprising six joints per arm and two grippers.
Platform-specific input and output adapters connect the common pretrained experts to these native interfaces.

\paragraph{Benefits of joint video--action pretraining.}
The paired WorldBridge results in Fig.~\ref{fig:real_robot_task_success_rates} show that omitting joint video--action pretraining reduces average success to 46.7\%, an absolute decrease of 25.8 percentage points.
This variant retains the pretrained Wan video backbone~\cite{wan2025wan}, so the comparison measures the additional benefit of our joint pretraining stage.
Pretraining improves all six tasks.
The largest gain occurs on sorting screws, where success increases from 10\% to 55\%; packing books and separating LEGO bricks each improve from 50\% to 80\%.
The screw-sorting sequence in Fig.~\ref{fig:real_robot_task_execution} illustrates the fine grasping and compartment placement required by the task with the largest pretraining gain.
On G2, the gains are 15 percentage points for arranging flowers, 20 for hanging cups, and 15 for scanning drinks.
The improvements therefore span precise small-object handling, spatial alignment, coordinated manipulation, and memory-dependent execution.
In particular, the substantial gains on the AC-1 tasks support the value of joint pretraining for downstream adaptation under the available task demonstrations.

\paragraph{Memory-dependent drink scanning.}
In this task, the robot must grasp a scanner, scan the original drinks in order, and finally scan an additional drink introduced by a human during execution.
Correct execution requires three aspects of memory.
First, the robot must remember each drink's placement destination after scanning, preserving the spatial association across grasping, scanning, and placement.
Second, it must track which drinks have already been scanned to select the next drink without repeating or omitting a scan.
Third, it must identify the newly introduced drink and defer its scan until the original drinks have been processed, preserving the order of task events despite the scene change.
The drink-scanning row in Fig.~\ref{fig:real_robot_task_execution} visualizes these dependencies through placement markers, highlighted scanned drinks, and the human introduction of a drink designated for the final scan.
A drink's current pose does not necessarily reveal its scanning status or when it entered the scene; visually similar observations can therefore demand different next actions depending on the execution history.
WorldBridge conditions action prediction on short- and long-term visual history, providing context for recent manipulation transitions and earlier task events.
As quantified in Fig.~\ref{fig:real_robot_task_success_rates}, WorldBridge achieves 65\% success, compared with 35\% for LingBot-VA and 5\% for $\pi_{0.5}$.
This advantage is consistent with visual memory helping resolve ambiguity about task state and maintain the required execution order during interaction.

\begin{table}[t]
  \centering
  \small
  \caption{Cumulative ablation of model-call latency for real-world deployment. Timings are measured after warmup with CUDA synchronization using BF16, recorded observations, precomputed T5 embeddings, and ten action denoising steps. Each configuration uses 60 calls across three measurement rounds. The checkpoint uses no historical observations. Timing includes conditioning VAE encoding and denoising, excluding JPEG processing, text encoding, communication, and robot execution.}
  \begin{tabular}{lrrr}
    \toprule
    Configuration & P50 (ms) & P95 (ms) & Speedup \\
    \midrule
    Joint video/action, 10/10 & 957.2 & 1085.0 & $1.00\times$ \\
    Decoupled video/action, 1/10 & 269.7 & 290.5 & $3.55\times$ \\
    + Inference mode and runtime caches & 218.7 & 260.0 & $4.38\times$ \\
    + QKV projection and KV buffers & 204.1 & 212.1 & $4.69\times$ \\
    + Attention switches (full eager profile) & 227.2 & 232.0 & $4.21\times$ \\
    + Compilation, without CUDA Graphs & 126.6 & 134.7 & $7.56\times$ \\
    + CUDA Graph replay & \textbf{103.9} & \textbf{105.6} & $\mathbf{9.21\times}$ \\
    \bottomrule
  \end{tabular}

  \label{tab:efficiency_development}
\end{table}

\paragraph{Latency ablation.} Table~\ref{tab:efficiency_development} reports a cumulative ablation of model-call latency. Reducing video denoising from ten steps to one while retaining ten action denoising steps lowers median latency from 957.2 to 269.7\,ms ($3.55\times$). Runtime caching and QKV/KV-buffer optimizations further reduce it to 204.1\,ms, while the full eager attention-switch profile raises it to 227.2\,ms. Compilation without CUDA Graphs lowers median latency to 126.6\,ms ($7.56\times$ speedup over baseline), and CUDA Graph replay reduces it further to 103.9\,ms ($9.21\times$), with a final P95 of 105.6\,ms. The explicit action FlashAttention switch does not improve eager latency, and some eager-path optimizations are superseded by compilation. With these optimizations, the deployed system sustains nearly 7\,Hz during real-world execution, including data processing and communication, while maintaining strong task success rates and demonstrating practical real-time usability.

\section{Related Work}

Vision-language-action (VLA) models use pretrained vision-language models to learn robot policies that follow language instructions. RT-2 transfers web knowledge to robot control~\cite{brohan2023rt2}, and OpenVLA provides an open-source model trained on diverse robot demonstrations~\cite{kim2024openvla}. $\pi_0$ uses flow matching to generate continuous actions~\cite{black2024pi0}, while $\pi_{0.5}$ uses heterogeneous training data to improve generalization across environments~\cite{physicalintelligence2025pi05}. However, these models still struggle with new tasks when action-labeled demonstrations are limited or unavailable. Combining action prediction with world modeling offers a way to learn both how tasks unfold and how to execute them. WorldVLA unifies image and action understanding and generation in an autoregressive model, showing that jointly learning the two can improve both action prediction and visual generation~\cite{cen2025worldvlaautoregressiveactionworld}. Video-based World Action Models (WAMs) further use video priors to support robot control beyond the coverage of action-labeled data. DreamZero shows that these priors can support unseen-task execution and transfer from video-only demonstrations~\cite{ye2026dreamzero}, while Zero-WAM uses human videos as in-context guidance for new tasks~\cite{zhou2026zerowam}. Other studies explore video-action pretraining~\cite{zhang2026native,wang2026openwam} and temporal context through longer observation histories or persistent memory~\cite{huang2026longwam,yang2026memorywam}. Our work examines how video coverage and action supervision contribute to task generalization, and how the video-to-action bridge turns predicted visual dynamics into executable actions.
\section{Conclusion}

We presented WorldBridge, a World Action Model that connects visual dynamics to executable control for task generalization. 
Our study highlights the complementary roles of video coverage and action grounding: additional task videos enable transfer beyond the tasks providing action supervision, while even a single action-labeled demonstration per new task substantially improves execution. 
Guided by these findings, WorldBridge couples video and action experts through structured attention and joint flow matching, incorporating temporal memory, action forcing, and unified pre-training on heterogeneous robot and egocentric demonstrations. 
Experiments across simulation benchmarks and two real-robot platforms demonstrate strong manipulation performance and the benefits of joint video–action pretraining. 
These results suggest that advancing task generalization requires both broader visual experience and a transferable video-to-action bridge. 
Extending this bridge to more diverse tasks with less action supervision remains an important direction for future work.
\section{Contributions}

\subsection*{Core Contributors}
Jun Cen\textsuperscript{5}, Hao Li\textsuperscript{2,3,4}, Liangkai Li\textsuperscript{1,4}, Yaoxu Lyu\textsuperscript{1,2,3,4}, Meng Wei\textsuperscript{2,3}, Zeqi Xiao\textsuperscript{3}, Sizhe Yang\textsuperscript{3,4}, Yifan Ye\textsuperscript{3,5}, Duo Zheng\textsuperscript{1,2,3}

\textit{Authors are listed in alphabetical order.} The superscripts denote: \textsuperscript{1} Data, \textsuperscript{2} Pretraining, \textsuperscript{3} Post-training, \textsuperscript{4} Real-world Infra and Deployment, \textsuperscript{5} Model Architecture.

\subsection*{Contributors}
Yubo Guo, Yuqiang Yang, Shanghang Zhang

\textit{Authors are listed in alphabetical order.} 

\subsection*{Supervision}
Jun Cen\textsuperscript{\ensuremath{\dagger}}

\par\vfill
\noindent\rule{\linewidth}{0.4pt}\par
\vspace{0.3em}
\noindent\textnormal{\textsuperscript{\ensuremath{\dagger}}\hspace{0.5em}Corresponding\ author: {cenjun.1@jd.com}}

\clearpage
\bibliographystyle{assets/plainnat}
\bibliography{references}

% \appendix
% \input{sections/99_appendix}

\end{document}